\documentclass[11pt]{article} % For LaTeX2e
\usepackage{jmlr2e}

\usepackage[utf8]{inputenc} % allow utf-8 input
\usepackage[T1]{fontenc}    % use 8-bit T1 fonts
\usepackage{hyperref}       % hyperlinks
\hypersetup{
    colorlinks=true,
    linkcolor=blue,
    citecolor=blue,
    urlcolor=blue,
}
\usepackage{enumitem} 
\usepackage{bm}
\usepackage{url}            % simple URL typesetting
\usepackage{wrapfig,lipsum,booktabs}       % professional-quality tables
\usepackage{amsfonts}       % blackboard math symbols
\usepackage{nicefrac}       % compact symbols for 1/2, etc.
\usepackage{microtype}      % microtypography

\usepackage{amssymb}
\usepackage[ruled,lined]{algorithm2e}
\usepackage{framed,graphicx}
\usepackage[lofdepth,lotdepth]{subfig}
\usepackage{color}
\usepackage{amsmath}
\usepackage{bigints}
\usepackage{multirow}
\newcommand{\argmin}{\operatornamewithlimits{argmin}}
\firstpageno{1}

\begin{document}

\title{\textsc{Apollo}: An Adaptive Parameter-wise Diagonal Quasi-Newton Method for Nonconvex Stochastic Optimization}
% Authors must not appear in the submitted version. They should be hidden
% as long as the \iclrfinalcopy macro remains commented out below.
% Non-anonymous submissions will be rejected without review.

\author{\name Xuezhe Ma\thanks{Work was done at Carnegie Mellon University.} \email xuezhema@isi.edu \\
\addr Information Sciences Institute \\
University of Southern California \\
%Los Angeles, CA, USA
}
\editor{}

\maketitle

\begin{abstract}
In this paper, we introduce \textsc{Apollo}, a quasi-Newton method for nonconvex stochastic optimization, which dynamically incorporates the curvature of the loss function by approximating the Hessian via a diagonal matrix.
Importantly, the update and storage of the diagonal approximation of Hessian is as efficient as adaptive first-order optimization methods with linear complexity for both time and memory.
To handle nonconvexity, we replace the Hessian with its rectified absolute value, which is guaranteed to be positive-definite.
Experiments on three tasks of vision and language show that \textsc{Apollo} achieves significant improvements over other stochastic optimization methods, including SGD and variants of Adam, in terms of both convergence speed and generalization performance.
The implementation of the algorithm is available at \url{https://github.com/XuezheMax/apollo}.
\end{abstract}

\section{Introduction}
Nonconvex stochastic optimization is of core practical importance in many fields of machine learning, in particular for training deep neural networks (DNNs).
First-order gradient-based optimization algorithms, conceptually attractive due to their linear efficiency on both the time and memory complexity, have led to tremendous progress and impressive successes. 
A number of advanced first-order algorithms have emerged over the years to pursue fast and stable convergence, among which stochastic gradient descent (SGD)~\citep{robbins1951stochastic,lecun1998gradient}, equipped with momentum~\citep{rumelhart1985learning,qian1999momentum,bottou2008tradeoffs}, has stood out for its simplicity and effectiveness across a wide range of applications~\citep{hinton2006reducing,hinton2012deep,graves2013generating}. 
However, one disadvantage of SGD is that the gradients in different directions are scaled uniformly, resulting in limited convergence speed and sensitive choice of learning rate, and thus has spawned a lot of recent interests in accelerating SGD from the algorithmic and practical perspectives.

Recently, many \emph{adaptive} first-order optimization methods have been proposed to achieve rapid training progress with element-wise scaled learning rates, and we can only mention a few here due to space limits.
In their pioneering work, \citet{duchi2011adaptive} proposed AdaGrad, which scales the gradient by the square root of the accumulative square gradients from the first iteration.
While AdaGrad works well for sparse settings, its performance significantly degrades for dense settings, primarily due to the monotonic increase of the accumulation.
Subsequently, several methods have been proposed with the intuition to limit the accumulation to a small window of past iterations, and in particular exponentially reduce the weight of earlier iterations.
Notable works incorporating this method are RMSProp~\citep{tieleman2012lecture}, AdaDelta~\citep{zeiler2012adadelta}, and Adam~\citep{kingma2015adam}, among which Adam has become the default optimization algorithm across many deep learning applications because of its fast convergence speed and relatively consistent selections of hyper-parameters~\citep{ruder2016overview,zhang2020adaptive}.
However, it has been observed that these adaptive optimization methods may converge to bad/suspicious local optima, resulting in worse generalization ability than their non-adaptive counterparts~\citep{wilson2017marginal}, or fail to converge due to unstable and extreme learning rates~\citep{luo2019adaptive}.

Quasi-Newton methods have been widely used in solving convex optimization problems, due to their efficient computation and fast convergence rate~\citep{broyden1967quasi,dennis1977quasi}.
However, the stochastic, high-dimensional and nonconvex nature of many machine learning tasks, such as training deep neural networks, has rendered many classical quasi-Newton methods ineffective and/or inefficient~\citep{keskar2016adaqn,wang2017stochastic,yao2020adahessian}.
Indeed, in many natural language processing (NLP) and computer vision (CV) tasks~\citep{he2016deep,ma-hovy-2016-end,luo2019adaptive}, SGD (with momentum) is chosen as the optimizer, benefiting from its stable and efficient training and outstanding generalization.

In this work, we develop \textsc{Apollo}, a quasi-Newton method for nonconvex stochastic optimization to simultaneously tackle the aforementioned challenges of stochastic variance, nonconvexity and inefficiency.
Algorithmically, \textsc{Apollo} dynamically incorporates the curvature of the objective function with diagonally approximated Hessian.
It only requires first-order gradients and updates the approximation of the Hessian diagonally so that it satisfies a parameter-wise version of the weak secant condition~\citep{wolfe1959secant}.
To handle nonconvexity, we replace the Hessian with its rectified absolute value, the computation of which is also efficient under our diagonal approximation, yielding an efficient optimization algorithm with linear complexity for both time and memory (\S\ref{sec:apollo}).
Experimentally, through three tasks on CV and NLP with popular deep neural networks, including ResNets~\citep{he2016deep}, LSTMs~\citep{hochreiter1997long} and Transformers~\citep{vaswani2017attention}, we demonstrate that \textsc{Apollo} significantly outperforms SGD and variants of Adam, in terms of both convergence speed and generalization performance (\S\ref{sec:experiments}).

\section{Backgrounds}
\label{sec:background}
In this section, we set up the notations on nonconvex stochastic optimization, briefly review the \mbox{(quasi-) Newton methods}, and discuss the problems of applying quasi-Newton methods to nonconvex stochastic optimization that we attempt to study in the rest of the paper.
\vspace{-1mm}
\subsection{Nonconvex Stochastic Optimization}
\label{subsec:problem}
In this paper, we consider the following nonconvex stochastic optimization problem:
\begin{equation}\label{eq:optimization}
\vspace{-2mm}
\min\limits_{\theta \in \mathcal{R}^d} f(\theta) = \mathbb{E} [l(\theta; \Gamma)]
\end{equation}
where $l: \mathcal{R}^d \times \mathcal{R}^n \rightarrow \mathcal{R}$ is a continuously differentiable (and possible nonconvex) function, $\theta \in \mathcal{R}^d$ denotes the parameter to be optimized, $\Gamma \in \mathcal{R}^n$ denotes a random variable with distribution function P, and $\mathbb{E}[\cdot]$ denotes the expectation w.r.t $\Gamma$.
Intuitively, $\Gamma$ incorporates noises in $f$, leading to a stochastic objective function.
A special case of \eqref{eq:optimization} that arises frequently in machine learning is the empirical risk minimization problem:
\begin{equation}\label{eq:erm}
\vspace{-2mm}
\min\limits_{\theta \in \mathcal{R}^d} f(\theta) = \frac{1}{N} \sum\limits_{i=1}^{N} l_{i}(\theta)
\end{equation}
where $l_i: \mathcal{R}^d \rightarrow \mathcal{R}$ is the loss function corresponding to the $i$-th data, and $N$ is the number of data samples that is assumed to be extremely large.
Objective functions may also have other sources of noise than data subsampling, such as dropout~\citep{srivastava2014dropout} in deep neural networks.
\vspace{-1mm}
\paragraph{Decoupled Parameters.} In this work, we consider a setting of decoupled parameters: $\theta = \{\theta^{(l)}, l = 1, 
\ldots, L\}$.
Intuitively, under this setting the parameter $\theta$ is decoupled into a sequence of parameters serving different functionalities.
For example, in neural network training the parameters of a neural network can be naturally decoupled into the parameters of different layers or modules.

\subsection{Newton and quasi-Newton Methods}
\label{subsec:newton}
Newton's method usually employs the following updates to solve \eqref{eq:optimization}:
\begin{equation}\label{eq:newton}
    \theta_{t+1} = \theta_{t} - H_t^{-1} g_t
\end{equation}
where $g_t = \nabla f(\theta_t)$ is the gradient at $\theta_t$ and $H_t = \nabla^2 f(\theta_t)$ is the Hessian matrix.
The convergence rate of Newton's method is \emph{quadratic} under standard assumptions~\citep{nocedal2006numerical}.
However, major challenges with this method are i) the expensive computation of the inverse Hessian at every iteration and the corresponding quadratic memory complexity; and ii) the limitation to convex functions (nonconvexity results in negative curvature of $H_t$ and misleads the update directions).

A standard alternative to Newton’s method is a class of quasi-Newton methods, which have been widely used in solving convex deterministic optimization problem:
\begin{equation}\label{eq:quasi-newton}
    \theta_{t+1} = \theta_{t} - \eta_t B_t^{-1} g_t
\end{equation}
where $\eta_t$ is the stepsize (a.k.a learning rate), $B_t$ is an approximation to the Hessian matrix $\nabla^2 f(\theta_t)$ at $\theta_t$, which is updated based on the well-known secant equation:
\begin{equation}\label{eq:secant}
\begin{array}{l}
    B_{t+1} = \argmin\limits_{B} \| B - B_t \| \\
    \textrm{s.t.} \quad B_{t+1} s_t = y_t \quad \textrm{(secant equation)}
\end{array}
\end{equation}
where $s_t = \theta_{t+1} - \theta_t$ and $y_t = g_{t+1} - g_t$.
$B_{t+1}$ is, in the sense of some matrix norm, the closest to $B_t$ among all symmetric matrices that satisfy the secant equation.
Each choice of the matrix norm results in a different update formula, such as DFP~\citep{davidon1991variable,fletcher1987practical} and BFGS~\citep{broyden1970convergence,fletcher1970new,goldfarb1970family,shanno1970conditioning}.
The popularity of this method is due to the fact that only the gradient of the objective function is required at each iteration. 
Since no second derivatives (Hessian) are required, quasi-Newton methods are sometimes more efficient than Newton's method, especially when the computation of Hessian is expensive.
To further reduce memory cost, one seminal work is the limited memory BFGS (L-BFGS)~\citep{liu1989limited,byrd1995limited} that achieves desirable linear computational and memory complexity by approximating the Hessian as a series of sum of first order information from previous iterations.

\subsection{Problems of quasi-Newton Methods}
Despite their impressive successes on convex deterministic optimization, quasi-Newton methods suffer from their own problems in more challenging scenarios.
In this section, we mainly discuss three problems preventing quasi-Newton methods from being applied to the scenario of large-scale nonconvex stochastic optimization.
Due to these problems, no quasi-Newton methods (to our best knowledge) designed for nonconvex optimization consistently outperform adaptive first-order algorithms w.r.t convergence speed and generalization performance.
The main goal of this work is to algorithmically design and experimentally demonstrate a novel quasi-Newton method, in hope of improving the convergence speed and generalization performance of nonconvex stochastic optimization eventually.
\vspace{-1mm}
\paragraph{Stochastic Variance.}
One challenge of quasi-Newton methods on nonconvex stochastic optimization \eqref{eq:optimization} is the variance introduced by the stochastic nature of the problem.
At each iteration, only the stochastic gradient $g_t$ is available, which is an unbiased estimation of the gradient $\nabla f(\theta_t)$ and may lead to an erroneous approximation of Hessian~\citep{byrd2011use}.
\vspace{-1mm}
\paragraph{Nonconvexity.}
Another key challenge in designing such quasi-Newton methods lies in the difficulty of preserving the positive-definiteness of $B_t$ in \eqref{eq:secant}, due to the nonconvexity of the objective function.
What is worse is that performing line search is infeasible in the stochastic setting, due to the presence of noise in the stochastic gradients~\citep{wang2017stochastic}.
\vspace{-1mm}
\paragraph{Computational and Memory Efficiency.}
Even though quasi-Newton methods are more efficient than Newton's method, the time and memory complexities are still relatively large compared with adaptive first-order methods.
For instance, L-BFGS requires to store first-order information from $m$ previous iterations with commonly $m\ge 5$, which is still too expensive for deep neural networks containing millions of parameters.
Moreover, adapting quasi-Newton methods to nonconvex stochastic optimization probably introduces additional computation, further slowing down these methods.

\section{Adaptive Parameter-Wise Diagonal Quasi-Newton}
\label{sec:apollo}
With the end goal of designing an efficient quasi-Newton method to solve the problem in \eqref{eq:optimization} in mind, we first propose to approximate the Hessian with a diagonal matrix, whose elements are determined by the variational approach subject to the \emph{parameter-wise} weak secant equation (\S\ref{subsec:diagonal}).
Then, we explain our stepsize bias correction technique to reduce the stochastic variance in \S\ref{subsec:bias}.
To handle nonconvexity, we directly use the rectified absolute value of the diagonally approximated Hessian as the preconditioning of the gradient (\S\ref{subsec:nonconvex}).
The initialization technique of \textsc{Apollo} allows us to eliminate one hyper-parameter (\S\ref{subsec:initialization}).
At last, we provide a theoretical analysis of \textsc{Apollo}'s convergence in both convex optimization and nonconvex stochastic optimization (\S\ref{subsec:convergence}).
The pseudo-code is shown in Algorithm~\ref{alg:apollo}.
\vspace{-1mm}
\subsection{Quasi-Newton Methods with Diagonal Hessian Approximation}
\label{subsec:diagonal}
As discussed in \citet{bordes2009sgd}, designing an efficient stochastic quasi-Newton algorithm involves a careful trade-off between the sparsity of the approximation matrix $B_t$ and the quality of its approximation of the Hessian $H_t$, and diagonal approximation is a reasonable choice~\citep{becker1988improving,zhu1999quasi}.
If $B$ is chosen to be a diagonal matrix satisfying \eqref{eq:secant}, one can obtain a formula similar to the SGD-QN algorithm~\citep{bordes2009sgd}.

An alternative of the secant equation in the updating formula \eqref{eq:secant}, as first introduced by \citet{nazareth1995if}, is the weak secant equation~\citep{dennis1993sizing}:
\begin{equation}\label{eq:weak-secant}
\begin{array}{l}
    \qquad B_{t+1} = \argmin\limits_{B} \| B - B_t \| \\
    \qquad \textrm{s.t.} \quad s_t^T B_{t+1} s_t = s_t^T y_t \quad \textrm{(weak secant equation)}
\end{array}
\end{equation}
The motivation of using the weak secant condition in diagonal quasi-Newton method is straight-forward: the standard mean-value theorem might not necessarily hold for vector-valued functions expressed in the secant equation, $B_{t+1} s_t = y_t \approx \nabla^2 f(\theta_t) s_t$.
Thus, we do not know whether there exists a vector $\tilde{\theta} \in \mathcal{R}^d$ such that $y_t = \nabla^2 f(\tilde{\theta}) s_t$~\citep{dennis1977quasi}.
On the other hand, the Taylor theorem ensures that there exists such $\tilde{\theta}$ that $s_t^T y_t = s_t^T \nabla^2 f(\tilde{\theta}) s_t$, leading to the reasonable assumption of the weak secant condition~\eqref{eq:weak-secant}.

Based on the variational technique proposed in \citet{zhu1999quasi}, the solution of \eqref{eq:weak-secant} with Frobenius norm is:
\begin{equation}\label{eq:update}
    \Lambda \triangleq B_{t+1} - B_t = \frac{s_t^T y_t - s_t^T B_{t} s_t}{\|s_t\|_4^4} \,\, \mathrm{Diag}(s_t^2)
\end{equation}
where $s_t^2$ is the element-wise square vector of $s_t$, $\mathrm{Diag}(s_t^2)$ is the diagonal matrix with diagonal elements from vector $s_t^2$, and $\|\cdot\|_4$ is the $4$-norm of a vector.

\paragraph{Parameter-Wise Weak Secant Condition.}
However, in optimization problems with high-dimensional parameter space, such as training deep neural networks with millions of parameters, the weak secant condition might be too flexible to produce a good Hessian approximation.
In the setting of decoupled parameters (\S\ref{subsec:problem}), we propose a parameter-wise version of the weak secant equation to achieve a trade-off between the secant and weak secant conditions: for each parameter $\theta^{(l)} \in \theta$, we update $B$ corresponding to $\theta^{(l)}$ by solving \eqref{eq:weak-secant} individually.
Remarkably, the secant condition restricts $B$ with an equation of a $d$-dimensional vector, while the weak secant condition relaxes it with a $1$-dimensional scalar.
The parameter-wise weak secant condition expresses the restriction as a $l$-dimension vector ($1 < l < d$), resulting in a reasonable trade-off.
The updating formula is the same as \eqref{eq:update} for each parameter-wise $B$.

\subsection{Stepsize Bias Correction}
\label{subsec:bias}
To mitigate the stochastic variance problem in stochastic quasi-Newton methods, \textsc{Apollo} utilizes stepsize bias correction on the stochastic gradients at each step $t$.
We know that the optimal stepsize $\eta_t$ equals to $1$ w.r.t the \emph{quadratic approximation} underlying Newton's method, if the Hessian approximation $B_t$ and the stochastic gradient $g_t$ are close to the exact Hessian $H_t$ and gradient $\nabla f(\theta_t)$, respectively. 
Inspired by this, we correct the stepsize bias in the stochastic gradient $g_t$ by replacing it with a corrected gradient $g_t' = \eta_t g_t$.
Together with the corresponding corrected $y_t' = g_{t+1}' - g_t' = \eta_t y_t$, we correct the updating term $\Lambda$ of $B_t$ in \eqref{eq:update} by replacing $y_t$ with $y_t'$:
\begin{equation}\label{eq:correct}
\Lambda' = \frac{s_t^T y_t' - s_t^T B_{t} s_t}{\|s_t\|_4^4} \,\, \mathrm{Diag}(s_t^2) = -\frac{d_t^T y_t + d_t^T B_{t} d_t}{\|d_t\|_4^4} \,\, \mathrm{Diag}(d_t^2)
\end{equation}
where $d_t = -s_t / \eta_t  = B_t^{-1} g_t$ is the corrected update direction. 
Note that after applying the step bias correction, the update formula of $B_t$ in \eqref{eq:correct} is independent with the stepsize $\eta_t$, eliminating the stepsize bias. 
Technically, the stepsize bias correction is designed to reduce the stochastic variance, rather than entirely discarding the stepsize. 
The \textsc{Apollo} algorithm (Algorithm~\ref{alg:apollo}) still incorporates the stepsize at every iteration to enforce convergence. 

\begin{algorithm}[t]
\SetAlgoLined
\DontPrintSemicolon
\SetKwInOut{Input}{Init}
\SetKwInOut{Output}{Output}
\newcommand\mycommfont[1]{\small\ttfamily{#1}}
\SetCommentSty{mycommfont}
\SetKwComment{Comment}{$\triangleright$\ }{}
%\KwResult{Write here the result }
\caption{\textsc{Apollo}, our proposed algorithm for nonconvex stochastic optimization. All operations on vectors are element-wise. Good default settings are $\beta=0.9$ and $\epsilon=1e^{-4}$.}
\textbf{Initial:} $m_0, d_0, B_0 \leftarrow 0, 0, 0$ \tcp*{Initialize $m_0, d_0, B_0$ to zero} 
\While{$t \in \{0, \ldots, T\}$}{
    \For{$\theta \in \{\theta^{1}, \ldots, \theta^{L}\}$}{
        $g_{t + 1} \leftarrow \nabla f_t(\theta_t)$ \tcp*{Calculate gradient at step $t$}
        $m_{t + 1} \leftarrow \frac{\beta (1 - \beta^t)}{1 - \beta^{t+1}} m_t + \frac{1 - \beta}{1 - \beta^{t+1}} g_{t + 1}$ \tcp*{Update bias-corrected moving average}
        $\alpha \leftarrow \frac{d_t^T (m_{t + 1} - m_t) + d_t^T B_{t} d_t}{(\| d_t \|_4 + \epsilon)^4}$ \tcp*{Calculate coefficient of $B$ update}
        $B_{t+1} \leftarrow B_t - \alpha \cdot \mathrm{Diag}(d_t^2) $ \tcp*{Update diagonal Hessian}
        $D_{t+1} \leftarrow \mathrm{rectify}(B_{t+1}, 0.01)$ \tcp*{Handle nonconvexity}
        $d_{t+1} \leftarrow D_{t+1}^{-1} m_{t+1}$ \tcp*{Calculate update direction}
        $\theta_{t+1} \leftarrow \theta_t - \eta_{t+1} d_{t+1}$ \tcp*{Update parameters}
    }
}
\Return{$\theta_T$}
\label{alg:apollo}
\end{algorithm}

Based on previous studies, incorporating exponential moving averages (EMVs) for the stochastic gradients significantly reduces the variance~\citep{kingma2015adam}.
We follow these works and apply EMV to $g_t$, together with the initialization bias correction:
\begin{equation}\label{eq:emv}
    m_{t + 1} = \frac{\beta (1 - \beta^t)}{1 - \beta^{t+1}} m_t + \frac{1 - \beta}{1 - \beta^{t+1}} g_{t + 1}
\end{equation}
where $0 < \beta < 1$ is the decay rate of EMV and $y_t$ in \eqref{eq:correct} is written as $m_{t+1} - m_t$.
Note that we do not apply moving average methods to the approximated Hessian, though the diagonal matrix is easier to be explicitly formed to average than full matrices.
Investigating the moving average of the diagonal $B_t$ might be an interesting direction of future work.

\subsection{Rectified Absolute Value of Hessian for Nonconvexity}
\label{subsec:nonconvex}
To guarantee convergence, quasi-Newton methods require the approximated Hessian matrix $B_t$ to be positive definite at each step.
The common strategy in previous studies is to solve the updating formula in \eqref{eq:secant} by restricting the candidate matrix $B$ to be symmetric positive definite. 
It is known that the BFGS update preserves the positive-definiteness of $B_{t+1}$ as long as the curvature condition $s_{t}^T y_t > 0$ holds, which can be guaranteed for strongly convex problem.
For nonconvex problem, the curvature condition can be satisfied by performing a line search, which is, however, expensive or even infeasible in stochastic setting, because the exact function values and gradient information are unavailable.
\citet{wang2017stochastic} proposed the stochastic damped L-BFGS (SdLBFGS) method that implicitly generates a positive definite matrix without line search.
However, it usually requires large history size ($m \ge 100$) to guarantee convergence, which is infeasible for large-scale optimization.

To handle nonconvexity, we adopt a different strategy that does not require the solution of $B_{t}$ in \eqref{eq:secant} to be positive definite.
Intuitively, we search for $B_t$ that is a good approximation of the real Hessian, which is not necessarily positive definite in nonconvex problem.
When we use $B_t$ as preconditioning to calculate the update direction, we use its absolute value:
\begin{equation}
    |B_t| = \sqrt{B_t^T B_t}
\end{equation}
where $\sqrt{\cdot}$ is the positive definite square root of a matrix.
The motivation of absolute value is straight-forward: for dimensions with large absolute values of curvature, the objective function could be very sharp and we would prefer to take relatively smaller steps than those flatter dimensions.
Since \textsc{Apollo} formulate $B_t$ as a diagonal matrix, the cost of computing $|B_t|$ is marginal.

\paragraph{Rectified Absolute Value of $B_t$}
For nonconvex objective functions, there exist inflection points whose curvatures are zero.
To prevent the steps from becoming arbitrarily large, we rectify the absolute value of $B_t$ with a convexity hyper-parameter $\sigma$:
\begin{equation}\label{eq:retify}
    D_t = \mathrm{rectify}(B_t, \sigma) = \max(|B_t|, \sigma)
\end{equation}
where the $\mathrm{rectify}(\cdot, \sigma)$ function is similar to the rectified linear unit (ReLU)~\citep{nair2010rectified} with threshold set to $\sigma$.
The update direction in \eqref{eq:correct} is then $d_t = D_t^{-1} m_t$.

AdaHessian~\citep{yao2020adahessian} used an idea similar to the absolute values of $B_t$ to handle nonconvexity, where the root mean square averaging is applied to compute the Hessian diagonal.
Different from \textsc{Apollo}, AdaHessian requires second-order information to compute the Hessian matvec oracle and approximate the Hessian diagonal using Hutchinson's method, which is significantly more costly.

\subsection{Initialization}
\label{subsec:initialization}
The rectified $D_t$ in \eqref{eq:retify} introduces one more hyper-parameter $\sigma$, limiting the application of \textsc{Apollo} in practice. 
In this section, we show that the zero initialization approach in \textsc{Apollo}, which initializes the moving average of gradient $m_0$, the parameter update direction $d_0$ and the diagonal approximation of Hessian $B_0$ as (vector of) zeros, leads to coupled stepsize $\eta$ and convexity $\sigma$, allowing us to eliminate one hyper-parameter of $\eta$ or $\sigma$.

\paragraph{Coupled Stepsize $\eta$ and Convexity $\sigma$. }
With the zero initialization of $m_0$, $d_0$ and $B_0$, the following theorem illustrates the relation between $\eta$ and $\sigma$ (details in Appendix~\ref{appendix:proof:thm1}):
\begin{theorem}\label{thm:coupled}
Given zero initialization of $m_0, \,\, d_0, \,\, \textrm{and} \,\, B_0$ and a fixed parameter intialization $\theta_0$. 
Suppose that we have two sets of hyper-parameters $\eta, \sigma$ and $\eta', \sigma'$ with the same ratio: $\frac{\eta}{\sigma} = \frac{\eta'}{\sigma'}$.
Then the convergence trajectories of these two sets of hyper-parameters are exactly the same:
\begin{equation}
    \theta_t = \theta_t', \,\, \forall t \in \{1, \ldots, T\}.
\end{equation}
where $\theta_t$ and $\theta_t'$ are the parameters of $(\eta, \sigma)$ and $(\eta', \sigma')$ at iteration $t$, respectively.
\end{theorem}
From Theorem~\ref{thm:coupled}, we observe that $\eta$ and $\sigma$ are coupled with each other and in practice we only need to tune one of them, leaving the other fixed.
Therefore, in our experiments (\S\ref{sec:experiments}), we fix $\sigma=0.01$ and tune $\eta$ on different problems\footnote{We changed $\sigma$ from 1 to 0.01 to make $\eta$ in a suitable range. See Appendix~\ref{appendix:sigma} for details.}.

\paragraph{Learning Rate Warmup for \textsc{Apollo}}
As discussed in \citet{kingma2015adam}, zero initialization leads to estimations biased towards zero in the initial iterations.
For the moving average $m_t$, this bias can be corrected by dividing the bias-correction term \eqref{eq:emv}.
For $d_t$ and $B_t$, however, we cannot derive such bias correction terms.
Fortunately, a simple linear warmup heuristic of $\eta$ at the beginning iterations achieves remarkably stable training.

\subsection{Convergence Analysis}
\label{subsec:convergence}
Similar to previous work~\citep{reddi2018convergence,chen2019convergence,zhuang2020adabelief}, we omit the initialization bias correction step, i.e. we use $m_{t} = \beta_t m_{t-1} + (1-\beta_t) g_{t}, \, 0 < \beta_t < 1, \, \forall t \in [T]$.

We first analyze the convergence of \textsc{Apollo} in convex optimization using the online learning framework~\citep{zinkevich2003online} for a sequence of convex cost functions $f_1(\theta), f_2(\theta), \ldots, f_T(\theta)$.
\begin{theorem}\label{thm:convegence:convex}
(Convergence in convex optimization)
Let $\{\theta_t\}$ be the sequence from \textsc{Apollo}.
Suppose $\eta_t = \frac{\eta}{\sqrt{t}}, \, 0 < \beta_t \le \beta \le 1 \, \| g_t\|_2 \le G, \frac{\| D_{t-1}\|_1}{\eta_{t-1}} \le \frac{\| D_t\|_1}{\eta_t}, \, \|\theta_t - \theta_{t'}\|_2 \le D, \, \forall t, t' \in [T]$.
For $\theta_t$ generated with the \textsc{Apollo} algorithm, we have the following bound on the regret:
\begin{equation}
R_T \le \frac{\sqrt{T} D^2 \| D_T\|_1}{2\eta(1 - \beta)} + \frac{\eta G^2}{1 - \beta} (2\sqrt{T} - 1) + \frac{D^2}{2(1 - \beta)}\sum\limits_{t=1}^{T} \frac{\beta_t^{2}}{\eta_t}
\end{equation}
\end{theorem}
The following result falls as an immediate corollary of the above result.
\begin{corollary}\label{col:convegence:convex}
Suppose $\beta_t = \beta \lambda^{t-1}, 0 < \lambda < 1$ in Theorem~\ref{thm:convegence:convex}, we have
\begin{equation}
R_T \le \frac{\sqrt{T} D^2 \| D_T\|_1}{2\eta(1 - \beta)} + \frac{\eta G^2}{1 - \beta} (2\sqrt{T} - 1) + \frac{D^2\beta^2}{2\eta(1 - \beta)(1 - \lambda^2)^2}
\end{equation}
\end{corollary}
Theorem~\ref{thm:convegence:convex} implies the regret of \textsc{Apollo} is upper bounded by $O(\sqrt{T})$ (proof in Appendix~\ref{appendix:convergence:convex}).
The conditions for Corollary~\ref{col:convegence:convex}, as in \citet{reddi2018convergence}, can be relaxed to $\beta_t = \beta / t$ and still ensures a regret of $O(\sqrt{T})$.

For nonconvex case, we analyze the convergence rate of \textsc{Apollo} with the similar derivations of that in \citet{chen2019convergence}, since \textsc{Apollo} belongs to the family of \emph{generalized Adam-type} methods (proof in Appendix~\ref{appendix:convergence:nonconvex}):
\begin{theorem}\label{thm:convegence:nonconvex}
(Convergence in nonconvex stochastic optimization) Under the assumptions: \\
$\bullet$ $f$ is lower bounded and differentiable; $\|\nabla f(\theta) - \nabla f(\theta') \|_2 \le L \| \theta - \theta'\|_2, \, \|D_t\|_{\infty} < L, \forall t, \, \theta, \, \theta'$. \\
$\bullet$ Both the true and stochastic gradient are bounded, i.e. $\| \nabla f(\theta_t)\|_2 \le H, \, \|g_t\|_2 \le H, \,\, \forall t$. \\
$\bullet$ Unbiased and independent noise in $g_t$, i.e. $g_t = \nabla f(\theta_t) + \zeta_t$, $\mathbb{E}[\zeta_t] = 0$, and $\zeta_i \perp \zeta_j, \,\, \forall i \neq j$. \\
Assume $\eta_t = \frac{\eta}{\sqrt{t}}, \, \beta_t \le \beta \le 1$ in non-increasing, $\frac{D_{t-1, j}}{\eta_{t-1}} \le \frac{D_{t, j}}{\eta_t}, \, \forall t \in [T], j \in [d]$, then:
\begin{equation}\label{eq:thm:nonconvex}
\min\limits_{t\in [T]} \mathbb{E}\left[ \| \nabla f(\theta_t) \|_2^2 \right] \le \frac{L}{\sqrt{T}}(C_1 \eta^2 H^2 (1 + \log T) + C_2 d\eta + C_3 d\eta^2 + C_4)
\end{equation}
\end{theorem}
where $C_1$, $C_2$, $C_3$ are constants independent of $d$ and $T$, $C_4$ is a constant independent of $T$, the expectation is taken w.r.t all the randomness corresponding to $\{ g_t\}$.
Theorem~\ref{thm:convegence:nonconvex} implies the convergence rate for \textsc{Apollo} in the non-convex case is $O(\log T / \sqrt{T})$, 
which is similar to Adam-type optimizer~\citep{reddi2018convergence,chen2019convergence}.
In addition, unlike Theorem 3.1 in \citet{chen2019convergence}, Theorem~\ref{thm:convegence:nonconvex} does not specify the bound of each update $\| \eta_t m_t / D_t\|_2$. 
This is because that, with conditions $\eta_t \le \eta$, $\|g_t\|_2 \le H$ and $D_t \ge 1$, it is straight-forward to derive the bound of $\| \eta_t m_t / D_t\|_2 \le \eta H = G$.

\section{Experiments}
\label{sec:experiments}
To evaluate \textsc{Apollo}, we conduct experiments on four benchmark datasets across three tasks of CV and NLP that are commonly used to evaluate optimization algorithms: CIFAR-10~\citep{krizhevsky2009learning} and ImageNet~\citep{deng2009imagenet} for image classification; One Billion Words~\citep{chelba2013one} for language modeling; and WMT 2014 English-German for neural machine translation.
The five baseline methods we compare with are SGD with momentum~\citep{bottou2008tradeoffs}, Adam~\citep{kingma2015adam}, Rectified Adam (RAdam)~\citep{liu2020variance}, AdaBelief~\citep{zhuang2020adabelief}, and AdaHessian~\citep{yao2020adahessian}.
Following \citet{loshchilov2019decoupled}, we decouple weight decays in Adam, RAdam, AdaBelief and AdaHessian in all the experiments\footnote{For AdaBelief, we also tried standard L$_2$ regularization.  But the accuracies are consistently worse than the models with decoupled weight decay.}. 
For each experiment, we report the average over 5 runs. More detailed descriptions, results and analysis of the conducted experiments are provided in Appendix~\ref{appendix:details}.

\begin{wraptable}{r}{0.58\textwidth}
\vspace{-5mm}
\caption{Test Acc. on CIFAR-10 and ImageNet.}
\label{tab:classify}
\centering
{\small\setlength{\tabcolsep}{4pt}\renewcommand{\arraystretch}{0.9}
\begin{tabular}[t]{l|cc|cc}
\toprule
 & \multicolumn{2}{c|}{\textbf{CIFAR-10}} & \multicolumn{2}{c}{\textbf{ImageNet}} \\
\textbf{Method} & \textbf{milestone} & \textbf{cosine} & \textbf{milestone} & \textbf{cosine} \\
\midrule
SGD   & 93.94 & 94.53 & 77.57 & 78.26 \\
\midrule
Adam  & 93.74 & 94.24 & 76.86 & 77.54 \\
RAdam & 93.88 & 94.38 & 76.91 & 77.68 \\
AdaBelief & 94.03 & 94.51 & 77.55 & 78.22 \\
\midrule
AdaHessian  & 93.97 & 94.48 & 77.61 & 78.02 \\
\midrule
\textbf{\textsc{Apollo}} & \textbf{94.21} & \textbf{94.64} & \textbf{77.85} & \textbf{78.45} \\
\bottomrule
\end{tabular}
}
\vspace{-3mm}
\end{wraptable}

\subsection{Image Classification}
\begin{figure}[t]
\centering
\includegraphics[width=1.0\textwidth]{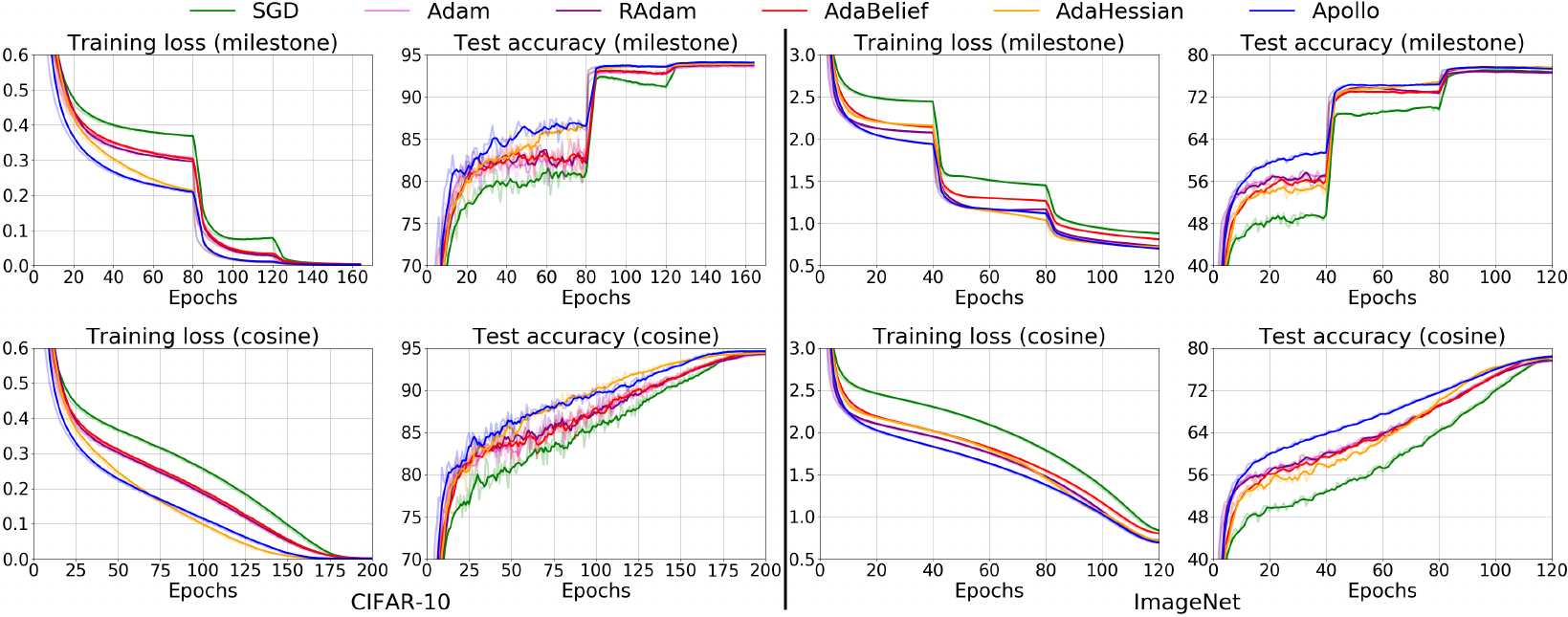}
\caption{Training loss and test accuracy of ResNet-110 on CIFAR-10 and ResNeXt-50 on ImageNet, with two schedule strategies of learning rate decay.}
\label{fig:classify}
\end{figure}

We begin our experiments with an evaluation of the convergence and generalization performance on image classification.
We use ResNet-110\footnote{ResNet-110 is a modified (small) version of ResNet-18 to adapt the image size $32\times 32$ in CIFAR-10.} for CIFAR-10 and standard ResNeXt-50~\citep{xie2017aggregated} for ImageNet, respectively.
The results on CIFAR-10 and ImageNet are presented in Figure~\ref{fig:classify} and Table~\ref{tab:classify}, together with the five baselines.
For each optimizer, we use two scheduling strategies of learning rate decay: i) milestone that decays the learning rate at the end of some predefined epochs; and ii) cosine annealing schedule proposed in \citet{loshchilov2017sgdr}.
All the optimization methods are comprehensively tuned, especially for the learning rate and the rate of weight decay.
It is because that the strength of weight decay regularization is co-related with the learning rate, even though the decoupled weight decay technique~\citep{loshchilov2019decoupled} has been applied.
The tuning information and the model details are provided in the Appendix~\ref{appendix:classification}.

From Figure~\ref{fig:classify} and Table~\ref{tab:classify}, we see that \textsc{Apollo} outperforms the four first-order methods (SGD, Adam, RAdam and AdaBelief) on both the convergence speed and classification accuracy, demonstrating its effectiveness on training the ResNet architectures based on convolutional neural networks (CNNs)~\citep{lecun1989backpropagation}.
Comparing with AdaHessian, \textsc{Apollo} obtains better test accuracy with similar convergence speed.
Note that AdaHessian requires second-order information and is significantly more costly (detailed comparison of time and memory costs in Appendix~\ref{appendix:cost}).
Thus, we omit AdaHessian from the following experiments in the rest of this paper.

\begin{figure}[t]
\centering
\includegraphics[width=1.0\textwidth]{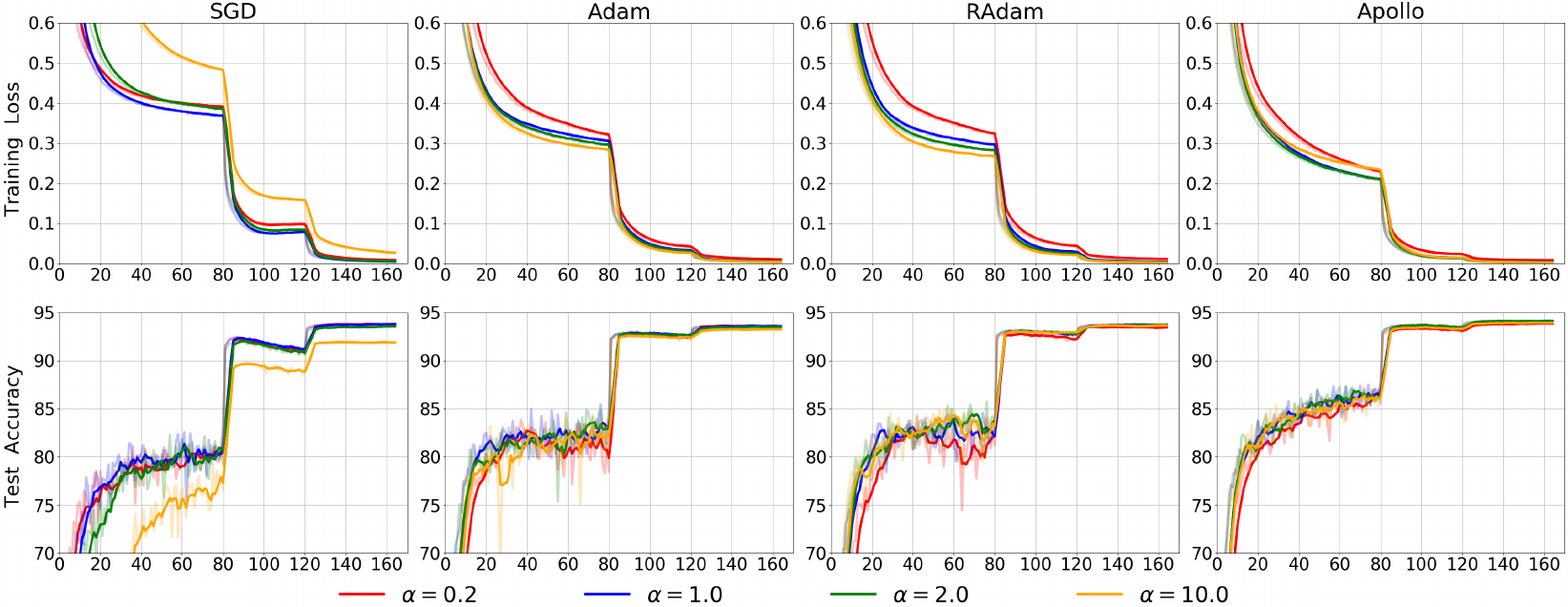}
\caption{SGD, Adam, RAdam and \textsc{Apollo} with different learning rates on CIFAR-10.}
\vspace{-2mm}
\label{fig:learning_rate}
\end{figure}

\paragraph{Robustness to Learning Rate Change.}
Besides performance improvements, we also investigate the robustness of different optimization methods to the change of learning rate.
For each optimizer, we use the learning rate in the previous experiment (Table~\ref{tab:classify}) as the base, i.e. $0.1$ for SGD, $0.001$ for Adam and RAdam, and $0.01$ for \textsc{Apollo}.
Then, we explore different learning rates that are $\alpha$ times of the base learning rate, with $\alpha \in \{0.2, 1.0, 2.0, 10.0\}$.
As mentioned above, we observed that the strength of weight decay regularization is co-related with the learning rate, even for Adam and RAdam with decoupled weight decay~\citep{loshchilov2019decoupled}.
To eliminate the impact of weight decay, we adjust the weight decay rates according to the factor $\alpha$.
Experimental results with ResNet-110 on CIFAR-10 are summarized in Figure~\ref{fig:learning_rate}.
After correcting the impact of weight decay, all the optimization methods, except SGD with $\alpha=10.0$, achieve consistent model performance (highly overlapped training and test curves with each other), while \textsc{Apollo} slightly improves the robustness of model training over the three baseline methods.
More detailed analysis on the effect of weight decay is provided in Appendix~\ref{appendix:weight_decay}.

\subsection{Language Modeling}
To evaluate \textsc{Apollo} on Recurrent Neural Networks (RNNs) that are applied in a wide range of NLP tasks~\citep{graves2013generating}, we conduct experiments on the One Billion Words dataset~\citep{chelba2013one}, using a two-layer LSTM network for language modeling (details in Appendix~\ref{appendix:lm}).

\begin{figure}[t]
\begin{minipage}{\textwidth}
\begin{minipage}[b]{0.75\textwidth}
\centering
\includegraphics[width=\textwidth]{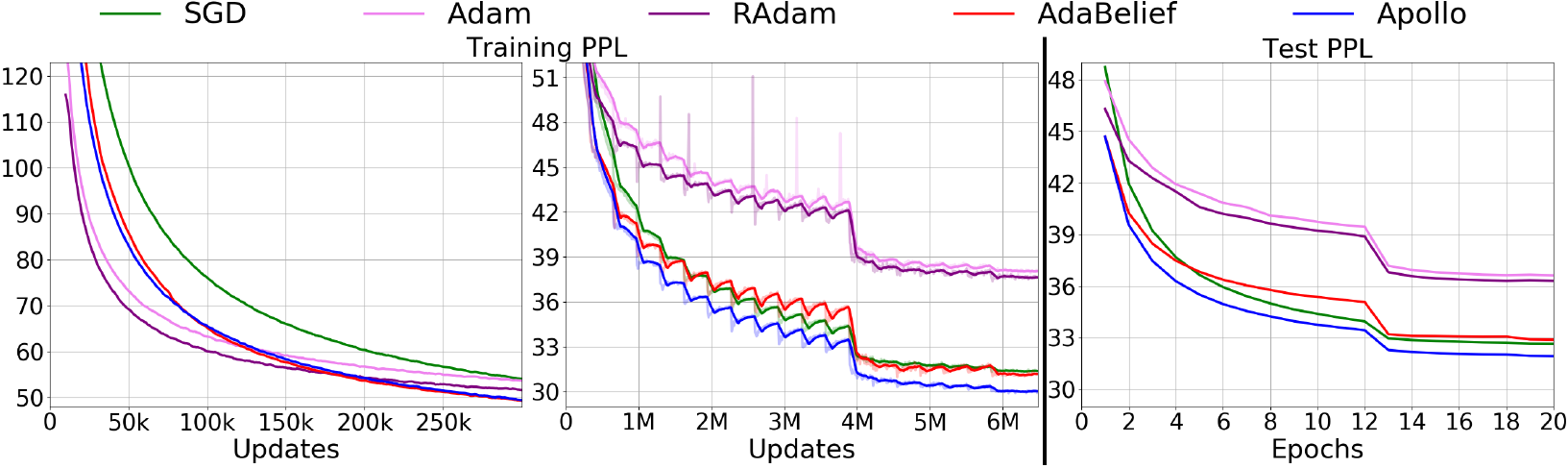}
\captionof{figure}{Language modeling (LSTMs) on One Billion Words.}
\label{fig:lm}
\end{minipage}
\hfill
\begin{minipage}[b]{0.24\textwidth}
\centering
\small{
\begin{tabular}[t]{l|c}
\toprule
\textbf{Method} & \textbf{PPL}\\
\midrule
SGD & 32.65 \\
\midrule
Adam & 36.68 \\
RAdam & 36.20 \\
AdaBelief & 32.83 \\
\midrule
\textbf{\textsc{Apollo}} & \textbf{31.94} \\
\bottomrule
\end{tabular}}
\captionof{table}{Test PPL.}
\label{tab:lm}
\end{minipage}
\end{minipage}
\end{figure}

Figure~\ref{fig:lm} and Table~\ref{tab:lm} illustrate the perplexity (PPL) of training and test for \textsc{Apollo} and four baseline methods, including SGD, Adam, RAdam and AdaBelief.
As shown in Figure~\ref{fig:lm}, although \textsc{Apollo} is slower than Adam-type methods in the first few updates, its convergence is much faster after that.
On generalization performance, \textsc{Apollo} achieves significant improvements (more than 4.0 PPL points on test data) over Adam and RAdam.
In addition, \textsc{Apollo} also outperforms AdaBelief, which obtains the lowest PPL among the three Adam-type optimization methods\footnote{We found that AdaBelief is very sensitive to the value of $\epsilon$. The result in Table~\ref{tab:lm} is obtained using $\epsilon=1e^{-12}$. With other values, e.g. $1e^{-8}$ or $1e^{-16}$, the PPL points of AdaBelief are even higher than Adam and RAdam. See Appendix~\ref{appendix:lm} for the details of hyper-parameter tuning.}.
This demonstrates the effectiveness of \textsc{Apollo} on training LSTM-based neural architectures. 

\paragraph{Training Stability. }
From the middle plot Figure~\ref{fig:lm} we see that the training losses of Adam and RAdam may suddenly increase.
This occurred in all the runs of experiments using Adam and RAdam, and we selected these successfully converged --- the loss went back to normal after some updates, and discarded those that failed to converge --- the model crashed due to loss numerical overflow.
The models optimized with \textsc{Apollo} never suffered from this issue, demonstrating the convergence stability of it.

\begin{wraptable}{r}{0.3\textwidth}
\vspace{-10mm}
\caption{Test BLEU.}
\label{tab:nmt}
\centering
\small{
\begin{tabular}[t]{l|c}
\toprule
\textbf{Method} & \textbf{BLEU} \\
\midrule
SGD         & 26.59$\pm0.07$ \\
\midrule
Adam        & 27.84$\pm0.12$ \\
RAdam       & 28.15$\pm0.15$ \\
AdaBelief   & 28.14$\pm0.11$ \\
\midrule
\textbf{\textsc{Apollo}} & \textbf{28.34}$\pm0.10$ \\
\bottomrule
\end{tabular}
}
\vspace{-3mm}
\end{wraptable}

\subsection{Neural Machine Translation}
To evaluate \textsc{Apollo} on Attention-based Transformer architecture~\citep{vaswani2017attention}, we train the Transformer-base model on the WMT2014 English-German (EN-DE) dataset (around 4.5M sentence pairs).
We use the same data preprocessing steps as in ~\citet{flowseq2019} (details in Appendix~\ref{appendix:nmt}).
We compare \textsc{Apollo} with the same four baseline methods in the experiments of language modeling.
For each experiment, we report the mean and standard variance over 5 runs.
From Table~\ref{tab:nmt}, the first interesting observation is that SGD performs much worse than Adam-type methods (which is opposite to its behaviour for ResNet- and LSTM-based neural architectures).
Similar observations about SGD were reported in \citep{yao2020adahessian,zhang2020adaptive}.
Despite this, \textsc{Apollo} obtains improvements over all the baseline methods for NMT with transformers.

\section{Related Work}
\paragraph{Stochastic Quasi-Newton Methods.}
In the literature of (nonconvex) stochastic quasi-Newton methods, several algorithms have been developed recently for large-scale machine learning problems: oLBFGS~\citep{schraudolph2007stochastic,mokhtari2015global}, RES~\citep{mokhtari2014res}, SFO~\citep{sohl2014fast}, SQN~\citep{byrd2016stochastic}, SdLBFGS~\citep{wang2017stochastic}, and AdaQN~\citep{keskar2016adaqn}, among which only SdLBFGS and AdaQN are designed to solve nonconvex optimization problems.
The SdLBFGS algorithm carefully controls the quality of modified BFGS updates to preserve the positive-definiteness of $B_t$ in \eqref{eq:secant} without line search.
AdaQN shares a similar idea but is specifically designed for RNNs by refining the initial L-BFGS scaling, step acceptance control and choice of curvature information matrix, and adopting the SQN framework~\citep{byrd2016stochastic}.
Different from these two methods, \textsc{Apollo} does not require $B_t$ in \eqref{eq:secant} to be positive definite, but replacing it with its rectified absolute value to handle nonconvexity.
Moreover, both SdLBFGS and AdaQN use the updating formula similar to L-BFGS and require even larger history size (commonly $\ge 100$) to guarantee convergence, preventing them from being applied to large-scale optimization. 
For comprehensive comparison of SdLBFGS with Apollo, we conducted experiments with small toy CNN models (details in Appendix~\ref{appendix:toy}).

\paragraph{Adaptive First-Order Methods.}
From the diagonal approximation of Hessian, \textsc{Apollo} is also related to those diagonally-scaled first-order algorithms, such as AdaGrad~\citep{duchi2011adaptive}, RMSProp~\citep{tieleman2012lecture}, AdaDelta~\citep{zeiler2012adadelta}, and Adam~\citep{kingma2015adam}.
Subsequently, a number of techniques have emerged to theoretically justify and algorithmically improve Adam, including AMSGrad~\citep{reddi2018convergence}, AdaBound~\citep{luo2019adaptive},  RAdam~\citep{liu2020variance} and AdaBelief~\citep{zhuang2020adabelief}.
The main difference is that the diagonal preconditioning in \textsc{Apollo} is directly derived from the quasi-Newton updating formula \eqref{eq:weak-secant}.
In terms of memory efficiency, \citet{anil2019memory} and \citet{chen2020extreme} further reduces the memory cost of adaptive methods, and \citet{agarwal2019efficient} proposed an efficient method for full-matrix adaptive regularization.

\paragraph{Stochastic Second-Order Hessian-Free Methods.}
Stochastic Second-Order Hessian-Free methods~\citep{martens2010deep,martens2011learning} implicitly solve quadratic models using matrix-vector products.
\citet{dauphin2014identifying} argued the existence of saddle points and proposed a method to rapidly escape them.
K-FAC~\citep{martens2015optimizing} computes a second-order step by constructing an invertible approximation of the Fisher information matrix in an online fashion. 
Shampoo~\citep{gupta2018shampoo} approximates the Fisher information matrix using low-rank decomposition.
Recently, \citet{yao2020adahessian} proposed AdaHessian, which approximates the Hessian diagonal using Hutchinson's method.
These second-order methods differ from \textsc{Apollo} mainly in the request of second-order information of the objective function at each iteration.

\section{Conclusion and Extensions}
We have introduced \textsc{Apollo}, a simple and computationally efficient quasi-Newton algorithm for nonconvex stochastic optimization.
This method is aimed towards large-scale optimization problems in the sense of large datasets and/or high-dimensional parameter spaces such as machine learning with deep neural networks.
Experimental results on three CV and NLP tasks demonstrate the effectiveness of \textsc{Apollo}, in terms of both convergence speed and generalization performance.
We briefly outline a few extensions to \textsc{Apollo} that we want to explore in future work.

\paragraph{Parameter-Wise Gradient Clipping.}
The standard gradient clipping method~\citep{pascanu2013difficulty} is to clip the gradients based on the norm computed over gradients of all the parameters together.
A modification of gradient clipping to properly apply it to \textsc{Apollo} is to clip the gradient of each parameter individually based on its own norm.
Preliminary results are provided in Appendix~\ref{appendix:pclip}.

\paragraph{Decoupled Weight Decay in \textsc{Apollo}.}
\citep{loshchilov2019decoupled} demonstrated that $L_2$ regularization is not identical to weight decay for adaptive gradient methods and proposed Adam with decoupled weight decay (AdamW).
The application of decoupled weight decay to \textsc{Apollo} is slightly different from AdamW as \textsc{Apollo} memorizes the update direction of the last iteration $d_t$ to update the diagonal Hessian. %$B_{t+1}$.
The algorithm of \textsc{Apollo} with decoupled weight decay is in Appendix~\ref{appendix:apollow}.

\paragraph{Making \textsc{Apollo} Scale-Invariant.}
An important advantage of adaptive optimization methods, including Adam and its variants, is that they are inherently scale-invariant --- invariant with the scale of the objective function.
The property of scale-invariance yields more consistent hyper-parameters of these adaptive methods than SGD across different machine learning tasks.
Unfortunately, \textsc{Apollo} does not hold the property of scale-invariance, and we need to ask if it is possible to make \textsc{Apollo} scale-invariant.
Interestingly, it is quite easy to develop a scale-invariant version of \textsc{Apollo} by applying a simple modification.
We provide more details about scale-invariant \textsc{Apollo} in Appendix~\ref{appendix:scale-invariant}.

\bibliography{apollo}

\newpage
\section*{Appendix: Apollo: An Adaptive Parameter-wise Diagonal Quasi-Newton Method for Nonconvex Stochastic Optimization}
\appendix
\section{Coupled Stepsize and Convexity}
\label{appendix:proof:thm1}
Before proving Theorem~\ref{thm:coupled}, we first define the notations.

Let $\alpha = \frac{\eta'}{\eta} = \frac{\sigma'}{\sigma}$ be the ratio of the two sets of learning rates.
Let $m'_t, \,\, d'_t$ and $B'_t$ be the corresponding terms of parameter $\theta'_t$ at step $t$ for $(\eta', \sigma')$.
\paragraph{Proof of Theorem 1}
\begin{proof}
Induction on the step of updates $t$, we attempt to prove that at each step $t$:
\begin{equation}\label{eq:statement}
    m_t = m'_t, \quad B'_t = \alpha{B_t}, \,\, \textrm{and} \,\, \theta_t = \theta'_t, \,\,\,  \forall t
\end{equation}
\paragraph{Initial step:} when $t=1$, since $\theta_0 = \theta'_0$, we have $m_1 = m'_1$.
With $d_0 = d'_0 = 0$ and \eqref{eq:correct}, we have $B_1 = B'_1 = 0$ and:
\begin{displaymath}
\begin{split}
D_1 & = \mathrm{rectify}(B_1, \sigma)  = \sigma \\
D'_1 & = \mathrm{rectify}(B'_1, \sigma')  = \sigma'
\end{split}
\end{displaymath}
Then, $D'_1 = \alpha D_1$ and 
\begin{displaymath}
\theta'_1 = \theta'_0 - \eta' D'^{-1}_1 m'_1 = \theta_0 - \eta \alpha (D^{-1}_1 / \alpha) m_1 = \theta_0 - \eta D^{-1}_1 m_1 = \theta_1.
\end{displaymath}
Thus, the statement \eqref{eq:statement} is true.
\paragraph{Induction on step $t$: } Suppose that the statement \eqref{eq:statement} is true for all the previous $t$ steps.
Now we prove the case $t+1$. 
From the inductive assumption and \eqref{eq:emv}, we have, 
\begin{displaymath}
B'_t = \alpha {B_t}, \quad d'_t = \frac{1}{\alpha}d_t \,\, \textrm{and} \,\, m_{t+1} = m'_{t+1}.
\end{displaymath}
From \eqref{eq:correct},
\begin{displaymath}
\begin{split}
B'_{t+1} & = B'_t - \frac{{d'}_t^T {y'}_t + {d'}_t^T B'_{t} d'_t}{\|d'_t\|_4^4} \,\, \mathrm{Diag}({d'}_t^2) \\
 & = \alpha{B_t} - \frac{(\frac{1}{\alpha} d_t)^T y_t + (\frac{1}{\alpha} d_t)^T (\alpha {B_t}) (\frac{1}{\alpha} d_t)}{\|(\frac{1}{\alpha} d_t)\|_4^4} \,\, \mathrm{Diag}((\frac{1}{\alpha} d_t)^2) \\
 & = \alpha {B_t} - \alpha \frac{d_t^T y_t + d_t^T B_t d_t}{\|d_t\|_4^4} \,\, \mathrm{Diag}(d_t^2) \\
 & = \alpha B_{t + 1}.
\end{split}
\end{displaymath}
Then, 
\begin{displaymath}
\begin{split}
D'_{t + 1} & = \mathrm{rectify}(B'_{t+1}, \sigma') \\
 & = \mathrm{rectify}(\alpha B_{t+1}, \alpha \sigma) \\
 & = \alpha \mathrm{rectify}(B_{t+1}, \sigma) \\
 & = \alpha D_{t+1}
\end{split}
\end{displaymath}
and we have $\theta'_{t+1} = \theta_{t+1}$.

Finally, to sum up with the induction, we have proven Theorem~\ref{thm:coupled}.
\end{proof}
\newpage

\section{Convergence Analysis}
\subsection{Convergence Analysis in Convex Optimization}
\label{appendix:convergence:convex}
\paragraph{Proof of Theorem 2}
\begin{proof}
Let $\theta^{*} = \argmin\limits_{\theta \in \mathcal{F}} \sum\limits_{t=1}^{T} f_t(\theta)$, where $\mathcal{F}$ is the feasible set of $\theta$.
As $\theta_{t+1} - \theta^{*} = \theta_{t} - \theta^{*} - \eta_t D_t^{-1} m_t$ and $m_t = \beta_t m_{t-1} + (1 - \beta_t) g_t$, we have the following:
\begin{displaymath}
\| D_t^{1/2}(\theta_{t+1} - \theta^{*})\|_2^{2} \le \| D_t^{1/2}(\theta_{t} - \theta^{*}) \|_2^{2} + \| \eta_t D_t^{-1/2} m_t \|_2^{2} - 2\eta_t (\beta_t m_{t-1} + (1 - \beta_t) g_t)^T (\theta_{t} - \theta^{*})
\end{displaymath}
Then, we have 
\begin{displaymath}
\begin{split}
g_{t}^T (\theta_{t} - \theta^{*}) & \le \frac{1}{2\eta_t(1 - \beta_t)} \left[ \| D_t^{1/2}(\theta_{t} - \theta^{*}) \|_2^{2} - \| D_t^{1/2}(\theta_{t+1} - \theta^{*}) \|_2^{2}\right] \\
 & \quad + \frac{\eta_t}{2 (1 - \beta_t)} \| D_t^{-1/2} m_t \|_2^{2} - \frac{\beta_t}{1 - \beta_t} m_{t-1}^T(\theta_{t} - \theta^{*}) \\
 & \le \frac{1}{2\eta_t(1 - \beta_t)} \left[ \| D_t^{1/2}(\theta_{t} - \theta^{*}) \|_2^{2} - \| D_t^{1/2}(\theta_{t+1} - \theta^{*}) \|_2^{2}\right] \\
 & \quad + \frac{\eta_t}{2 (1 - \beta_t)} \| D_t^{-1/2} m_t \|_2^{2} + \frac{\eta_t}{2(1 - \beta_t)} \| m_{t-1}\|_2^2 + \frac{\beta_t^2}{2\eta_t(1 - \beta_t)} \| \theta_{t} - \theta^{*}\|_2^2
\end{split}
\end{displaymath}
Using the standard approach of bounding the regret at each step with convexity of the functions $\{f_t\}_{t=1}^{T}$, we have the following bound of $R_T = \sum\limits_{t=1}^{T} f_t(\theta_t) - f_t(\theta^{*})$:
\begin{equation}\label{eq:thm2:pf1}
\begin{split}
\sum\limits_{t=1}^{T} f_t(\theta_t) - f_t(\theta^{*}) & \le \sum\limits_{t=1}^{T} g_{t}^T (\theta_{t} - \theta^{*}) \\
 & \le \sum\limits_{t=1}^{T} \frac{1}{2\eta_t(1 - \beta_t)} \left[ \| D_t^{1/2}(\theta_{t} - \theta^{*}) \|_2^{2} - \| D_t^{1/2}(\theta_{t+1} - \theta^{*}) \|_2^{2}\right] \\
 & + \sum\limits_{t=1}^{T} \frac{\eta_t}{2 (1 - \beta_t)} \| D_t^{-1/2} m_t \|_2^{2} + \frac{\eta_t}{2(1 - \beta_t)} \| m_{t-1}\|_2^2 \\
 & + \sum\limits_{t=1}^{T} \frac{\beta_t^2}{2\eta_t(1 - \beta_t)} \| \theta_{t} - \theta^{*}\|_2^2
\end{split}
\end{equation}
As $\| \theta_{t} - \theta^{*}\|_2 \le D$, $\beta_t < \beta < 1$ and $\|D_t\|_1 / \eta_t \ge \|D_{t-1}\|_1/\eta_{t-1}$, we have
\begin{equation}\label{eq:thm2:pf2}
\begin{split}
 & \sum\limits_{t=1}^{T} \frac{1}{2\eta_t(1 - \beta_t)} \left[ \| D_t^{1/2}(\theta_{t} - \theta^{*}) \|_2^{2} - \| D_t^{1/2}(\theta_{t+1} - \theta^{*}) \|_2^{2}\right] \\
= & \frac{\| D_1^{1/2}(\theta_{1} - \theta^{*}) \|_2^{2}}{2\eta_1(1 - \beta_1)} + \sum\limits_{t=2}^{T} \left[ \frac{\| D_t^{1/2}(\theta_{t} - \theta^{*}) \|_2^{2}}{2\eta_t(1 - \beta_t)} - \frac{\| D_{t-1}^{1/2}(\theta_{t} - \theta^{*}) \|_2^{2}}{2\eta_{t-1} (1 - \beta_{t-1})}\right] \\
\le & \frac{\| D_1^{1/2}(\theta_{1} - \theta^{*}) \|_2^{2}}{2\eta_1(1 - \beta)} + \frac{1}{2(1 - \beta)}\sum\limits_{t=2}^{T} \left[ \frac{\| D_t^{1/2}(\theta_{t} - \theta^{*}) \|_2^{2}}{\eta_t} - \frac{\| D_{t-1}^{1/2}(\theta_{t} - \theta^{*}) \|_2^{2}}{\eta_{t-1}} \right] \\
\le & \frac{\|(\theta_{1} - \theta^{*}) \|_2^{2}}{2\eta_1(1 - \beta)}\| D_1^{1/2}\|_2^{2} + \frac{1}{2(1 - \beta)}\sum\limits_{t=2}^{T} \|(\theta_{t} - \theta^{*})\|_2^2\left[ \frac{\| D_t^{1/2} \|_2^{2}}{\eta_t} - \frac{\| D_{t-1}^{1/2} \|_2^{2}}{\eta_{t-1}} \right]
\end{split}
\end{equation}
Since $\|D_t^{1/2}\|_2^2 = \| D_t\|_1$, we can rewrite the RHS of \eqref{eq:thm2:pf2} as:
\begin{equation}\label{eq:thm2:pf2_2}
\begin{split}
 & \frac{\|(\theta_{1} - \theta^{*}) \|_2^{2}}{2\eta_1(1 - \beta)}\| D_1^{1/2}\|_2^{2} + \frac{1}{2(1 - \beta)}\sum\limits_{t=2}^{T} \|(\theta_{t} - \theta^{*})\|_2^2\left[ \frac{\| D_t^{1/2} \|_2^{2}}{\eta_t} - \frac{\| D_{t-1}^{1/2} \|_2^{2}}{\eta_{t-1}} \right] \\
= & \frac{\|(\theta_{1} - \theta^{*}) \|_2^{2}}{2\eta_1(1 - \beta)}\| D_1\|_1 + \frac{1}{2(1 - \beta)}\sum\limits_{t=2}^{T} \|(\theta_{t} - \theta^{*})\|_2^2\left[ \frac{\| D_t \|_1}{\eta_t} - \frac{\| D_{t-1} \|_1}{\eta_{t-1}} \right] \\
\le & \frac{D^{2}}{2\eta_1(1 - \beta)}\| D_1\|_1 + \frac{D^2}{2(1 - \beta)}\sum\limits_{t=2}^{T} \left[ \frac{\| D_t \|_1}{\eta_t} - \frac{\| D_{t-1} \|_1}{\eta_{t-1}} \right] \\
= & \frac{D^2 \| D_T\|_1}{2\eta_T(1 - \beta)} = \frac{\sqrt{T} D^2 \| D_T\|_1}{2\eta(1 - \beta)}
\end{split}
\end{equation}
To sum up with \eqref{eq:thm2:pf1} and \eqref{eq:thm2:pf2_2}, we have
\begin{equation}\label{eq:thm2:pf3}
\begin{split}
R_T = \sum\limits_{t=1}^{T} f_t(\theta_t) - f_t(\theta^{*}) & \le \frac{\sqrt{T} D^2 \| D_T\|_1}{2\eta(1 - \beta)} \\ 
 & + \sum\limits_{t=1}^{T} \frac{\eta_t}{2 (1 - \beta_t)} \| D_t^{-1/2} m_t \|_2^{2} + \frac{\eta_t}{2(1 - \beta_t)} \| m_{t-1}\|_2^2 \\
 & + \sum\limits_{t=1}^{T} \frac{\beta_t^2}{2\eta_t(1 - \beta_t)} \| \theta_{t} - \theta^{*}\|_2^2 
\end{split}
\end{equation}
Since the element of $D_t$ is rectified by 1, i.e. $D_{t,i} \ge 1$, and $\| m_t\|_2 \le G$, $\beta_t < \beta < 1$, we have
\begin{equation}\label{eq:thm2:pf4}
\begin{split}
\sum\limits_{t=1}^{T} \frac{\eta_t}{2(1 - \beta_t)} \| D_t^{-1/2} m_t \|_2^{2} + \frac{\eta_t}{2(1 - \beta_t)} \| m_{t-1}\|_2^2 & \le \sum\limits_{t=1}^{T} \frac{\eta_t}{2(1 - \beta_t)} \| m_t \|_2^{2} + \frac{\eta_t}{2(1 - \beta_t)} \| m_{t-1}\|_2^2 \\
 & \le \frac{G^2}{1 - \beta} \sum\limits_{t=1}^{T} \eta_t = \frac{\eta G^2}{1 - \beta} \sum\limits_{t=1}^{T} \frac{1}{\sqrt{t}} \\
 & \le \frac{\eta G^2}{1 - \beta} (2 \sqrt{T} - 1)
\end{split}
\end{equation}
The last inequality is due to the following upper bound:
\begin{displaymath}
\sum\limits_{t=1}^{T} \frac{1}{\sqrt{t}} \le \int_{t=1}^{T} \frac{\mathrm{d}t}{\sqrt{t}} = 2 \sqrt{T} - 1
\end{displaymath}
Again, as $\| \theta_{t} - \theta^{*}\|_2 \le D$ and $\beta_t < \beta < 1$, we have
\begin{equation}\label{eq:thm2:pf5}
\sum\limits_{t=1}^{T} \frac{\beta_t^2}{2\eta_t(1 - \beta_t)} \| \theta_{t} - \theta^{*}\|_2^2 \le \frac{D^2}{2(1 - \beta)}\sum\limits_{t=1}^{T} \frac{\beta_t^2}{\eta_t}
\end{equation}
Finally, to sum up with \eqref{eq:thm2:pf3}, \eqref{eq:thm2:pf4} and \eqref{eq:thm2:pf5}, we have 
\begin{displaymath}
R_T \le \frac{\sqrt{T} D^2 \| D_T\|_1}{2\eta(1 - \beta)} + \frac{\eta G^2}{1 - \beta} (2\sqrt{T} - 1) + \frac{D^2}{2(1 - \beta)}\sum\limits_{t=1}^{T} \frac{\beta_t^{2}}{\eta_t}
\end{displaymath}
\end{proof}
\paragraph{Proof of Corollary 2.1}
\begin{proof}
Since $\beta_t = \beta \lambda^{t-1}$, by sum of arithmetico-geometric series we have
\begin{equation}\label{eq:cor2}
\sum\limits_{t=1}^{T} \lambda^{2(t-1)} \sqrt{t} \le \sum\limits_{t=1}^{T} \lambda^{2(t-1)} t \le \frac{1}{(1 - \lambda^2)^2}
\end{equation}
Plugging \eqref{eq:cor2} into \eqref{eq:thm2:pf5}, we have
\begin{displaymath}
R_T \le \frac{\sqrt{T} D^2 \| D_T\|_1}{2\eta(1 - \beta)} + \frac{\eta G^2}{1 - \beta} (2\sqrt{T} - 1) + \frac{D^2\beta^2}{2\eta(1 - \beta)(1 - \lambda^2)^2}.
\end{displaymath}
\end{proof}

\subsection{Convergence Analysis in Nonconvex Stochastic Optimization}
\label{appendix:convergence:nonconvex}
To prove Theorem~\ref{thm:convegence:nonconvex} in \S\ref{subsec:convergence}, we first describe the Theorem 3.1 in \citep{chen2019convergence}:
\paragraph{Theorem 3.1}~\citep{chen2019convergence} For an Adam-type method under the assumptions: \\
$\bullet$ $f$ is lower bounded and differentiable; $\|\nabla f(\theta) - \nabla f(\theta') \|_2 \le L \| \theta - \theta'\|_2, \, \forall \theta, \, \theta'$. \\
$\bullet$ Both the true and stochastic gradient are bounded, i.e. $\| \nabla f(\theta_t)\|_2 \le H, \, \|g_t\|_2 \le H, \,\, \forall t$. \\
$\bullet$ Unbiased and independent noise in $g_t$, i.e. $g_t = \nabla f(\theta_t) + \zeta_t$, $\mathbb{E}[\zeta_t] = 0$, and $\zeta_i \perp \zeta_j, \,\, \forall i \neq j$. \\
Assume $\beta_t \le \beta \le 1$ in non-increasing, $\| \eta_t m_t / \sqrt{v_t}\|_2 \le G$, then:
\begin{align}
& \mathbb{E}\left[ \sum\limits_{t=1}^{T} \eta_t \langle \nabla f(\theta_t), \nabla f(\theta_t)/\sqrt{v_t} \rangle \right] \nonumber \\
\le & \mathbb{E}\left[ C_1 \sum\limits_{t=1}^{T} \left\| \frac{\eta_t g_t}{\sqrt{v_t}}\right\|_2^{2} + C_2 \sum\limits_{t=2}^{T} \left\| \frac{\eta_t}{\sqrt{v_t}} - \frac{\eta_{t-1}}{\sqrt{v_{t-1}}} \right\|_1 + C_3 \sum\limits_{t=2}^{T-1} \left\| \frac{\eta_t}{\sqrt{v_t}} - \frac{\eta_{t-1}}{\sqrt{v_{t-1}}} \right\|_2^2\right] + C_4
\end{align}
where $C_1$, $C_2$, $C_3$ are constants independent of $d$ and $T$, $C_4$ is a constant independent of $T$, the expectation is taken w.r.t all the randomness corresponding to $\{ g_t\}$.

Since \textsc{Apollo} belongs to the family of \emph{generalized Adam}~\citep{chen2019convergence} with $\sqrt{v_t}$ corresponding to $D_t$, we have 
\begin{align}\label{eq:thm3:pf0}
& \mathbb{E}\left[ \sum\limits_{t=1}^{T} \eta_t \langle \nabla f(\theta_t), \nabla f(\theta_t)/D_t \rangle \right] \nonumber \\
\le & \mathbb{E}\left[ C_1 \sum\limits_{t=1}^{T} \left\| \frac{\eta_t g_t}{D_t}\right\|_2^{2} + C_2 \sum\limits_{t=2}^{T} \left\| \frac{\eta_t}{D_t} - \frac{\eta_{t-1}}{D_{t-1}} \right\|_1 + C_3 \sum\limits_{t=2}^{T-1} \left\| \frac{\eta_t}{D_t} - \frac{\eta_{t-1}}{D_{t-1}} \right\|_2^2\right] + C_4
\end{align}
Note that \textsc{Apollo} does not specify the bound of each update $\| \eta_t m_t / D_t\|_2$, because it is straight-forward to derive the bound with conditions $\eta_t \le \eta$, $\|g_t\|_2 \le H$ and $D_t \ge 1$.
\paragraph{Proof of Theorem 3}
With \eqref{eq:thm3:pf0}, we can prove our Theorem~\ref{thm:convegence:nonconvex} with similar derivations in \citet{chen2019convergence}.
\begin{proof}
We first bound non-constant terms in RHS of \eqref{eq:thm3:pf0}, which is given by
\begin{displaymath}
\mathbb{E}\left[ C_1 \sum\limits_{t=1}^{T} \left\| \frac{\eta_t g_t}{D_t}\right\|_2^{2} + C_2 \sum\limits_{t=2}^{T} \left\| \frac{\eta_t}{D_t} - \frac{\eta_{t-1}}{D_{t-1}} \right\|_1 + C_3 \sum\limits_{t=2}^{T-1} \left\| \frac{\eta_t}{D_t} - \frac{\eta_{t-1}}{D_{t-1}} \right\|_2^2\right]
\end{displaymath}
For the term with $C_1$, since $D_t \ge 1$, we have
\begin{equation}\label{eq:thm3:pf1}
\begin{split}
\mathbb{E}\left[\sum\limits_{t=1}^{T} \left\| \frac{\eta_t g_t}{D_t}\right\|_2^{2} \right] & \le \mathbb{E}\left[\sum\limits_{t=1}^{T} \left\| \eta_t g_t \right\|_2^{2} \right] \\
 & = \mathbb{E}\left[\sum\limits_{t=1}^{T} \left\| \left( \frac{\eta}{\sqrt{t}}\right) g_t \right\|_2^{2} \right] \\
 & \le \eta^2 H^2 \sum\limits_{t=1}^{T} \frac{1}{t} \le \eta^2 H^2 (1 + \log T)
\end{split}
\end{equation}
where the last inequality is due to $\sum_{t=1}^{T} 1/ t \le 1 + \log T$. \\
For the term with $C_2$, we have
\begin{equation}\label{eq:thm3:pf2}
\begin{split}
\mathbb{E}\left[ \sum\limits_{t=2}^{T} \left\| \frac{\eta_t}{D_t} - \frac{\eta_{t-1}}{D_{t-1}} \right\|_1 \right] & = \mathbb{E}\left[ \sum\limits_{j=1}^{d}\sum\limits_{t=2}^{T} \left( \frac{\eta_{t-1}}{D_{t-1, j}} - \frac{\eta_{t}}{D_{t, j}} \right)\right] \\
 & = \mathbb{E}\left[ \sum\limits_{j=1}^{d}\left( \frac{\eta_{1}}{D_{1, j}} - \frac{\eta_{T}}{D_{T, j}} \right) \right] = \mathbb{E}\left[ \sum\limits_{j=1}^{d}\frac{\eta_{1}}{D_{1, j}}\right] \le d\eta
\end{split}
\end{equation}
where the first equality is due to $\frac{D_{t-1, j}}{\eta_{t-1}} \le \frac{D_{t, j}}{\eta_t}, \, \forall t \in [T], j \in [d]$ and the second equality is due to telescope sum.\\
For the term with $C_3$, we have 
\begin{equation}\label{eq:thm3:pf3}
\mathbb{E}\left[\sum\limits_{t=2}^{T-1} \left\| \frac{\eta_t}{D_t} - \frac{\eta_{t-1}}{D_{t-1}} \right\|_2^2\right] \le \mathbb{E}\left[\eta \sum\limits_{t=2}^{T-1} \left\| \frac{\eta_t}{D_t} - \frac{\eta_{t-1}}{D_{t-1}} \right\|_1\right] \le d\eta^2
\end{equation}
where the first inequality is due to $|\eta_{t-1}/D_{t-1, j} - \eta_t/D_{t, j}| \le \eta$. \\
Then for \textsc{Apollo} we have 
\begin{equation}\label{eq:thm3:pf4}
\begin{split}
 & \mathbb{E}\left[ C_1 \sum\limits_{t=1}^{T} \left\| \frac{\eta_t g_t}{D_t}\right\|_2^{2} + C_2 \sum\limits_{t=2}^{T} \left\| \frac{\eta_t}{D_t} - \frac{\eta_{t-1}}{D_{t-1}} \right\|_1 + C_3 \sum\limits_{t=2}^{T-1} \left\| \frac{\eta_t}{D_t} - \frac{\eta_{t-1}}{D_{t-1}} \right\|_2^2\right] + C_4 \\
\le & C_1 \eta^2 H^2 (1 + \log T) + C_2 d\eta + C_3 d\eta^2 + C_4 
\end{split}
\end{equation}
Now we lower bound the LHS of \eqref{eq:thm3:pf0}. With the assumption $\|D_t\|_{\infty} \le L$, we have 
\begin{displaymath}
(\eta_t / D_t)_j \ge \frac{\eta}{L\sqrt{t}}
\end{displaymath}
And thus 
\begin{equation}\label{eq:thm3:pf5}
\mathbb{E}\left[ \sum\limits_{t=1}^{T} \eta_t \langle \nabla f(\theta_t), \nabla f(\theta_t)/D_t \rangle \right] \ge \mathbb{E}\left[ \sum\limits_{t=1}^{T} \frac{\eta}{L\sqrt{t}} \| \nabla f(\theta_t)\|_{2}^{2} \right] \ge \frac{\sqrt{T}}{L} \min\limits_{t\in [T]} \mathbb{E}\left[ \| \nabla f(\theta_t) \|_2^2 \right]
\end{equation}
Then, to sum up with \eqref{eq:thm3:pf0}, \eqref{eq:thm3:pf4} and \eqref{eq:thm3:pf5}, we have 
\begin{displaymath}
\frac{\sqrt{T}}{L} \min\limits_{t\in [T]} \mathbb{E}\left[ \| \nabla f(\theta_t) \|_2^2 \right] \le C_1 \eta^2 H^2 (1 + \log T) + C_2 d\eta + C_3 d\eta^2 + C_4 
\end{displaymath}
which is equivalent to
\begin{displaymath}
\begin{split}
\min\limits_{t\in [T]} \mathbb{E}\left[ \| \nabla f(\theta_t) \|_2^2 \right] & \le \frac{L}{\sqrt{T}}(C_1 \eta^2 H^2 (1 + \log T) + C_2 d\eta + C_3 d\eta^2 + C_4) \\
 & = \frac{1}{\sqrt{T}} (Q_1 + Q_2\log T)
\end{split}
\end{displaymath}
This completes the proof.
\end{proof}

\section{\textsc{Apollo} with Decoupled Weight Decay}
\label{appendix:apollow}
\begin{algorithm}[h]
\SetAlgoLined
\DontPrintSemicolon
\SetKwInOut{Input}{Init}
\SetKwInOut{Output}{Output}
\newcommand\mycommfont[1]{\small\ttfamily{#1}}
\SetCommentSty{mycommfont}
\SetKwComment{Comment}{$\triangleright$\ }{}
%\KwResult{Write here the result }
\caption{\textsc{Apollo} with weight decay (\textcolor{blue}{$L_2$}/\textcolor{red}{Decoupled})}
\textbf{Initial:} $m_0, d_0, B_0 \leftarrow 0, 0, 0$ \tcp*{Initialize $m_0, d_0, B_0$ to zero} 
\While{$t \in \{0, \ldots, T\}$}{
    \For{$\theta \in \{\theta^{1}, \ldots, \theta^{L}\}$}{
        $g_{t + 1} \leftarrow \nabla f_t(\theta_t) \textcolor{blue}{+ \gamma \theta_t}$ \tcp*{Calculate gradient at step $t$}
        $m_{t + 1} \leftarrow \frac{\beta (1 - \beta^t)}{1 - \beta^{t+1}} m_t + \frac{1 - \beta}{1 - \beta^{t+1}} g_{t + 1}$ \tcp*{Update bias-corrected moving}
        $\alpha \leftarrow \frac{d_t^T (m_{t + 1} - m_t) + d_t^T B_{t} d_t}{(\| d_t \|_4 + \epsilon)^4}$ \tcp*{Calculate coefficient of $B$ update}
        $B_{t+1} \leftarrow B_t - \alpha \cdot \mathrm{Diag}(d_t^2) $ \tcp*{Update diagonal Hessian}
        $D_{t+1} \leftarrow \mathrm{rectify}(B_{t+1}, 0.01)$ \tcp*{Handle nonconvexity}
        $d_{t+1} \leftarrow D_{t+1}^{-1} m_{t+1} \textcolor{red}{+ \gamma \theta_t}$ \tcp*{Calculate update direction}
        $\theta_{t+1} \leftarrow \theta_t - \eta_{t+1} d_{t+1}$ \tcp*{Update parameters}
    }
}
\Return{$\theta_T$}
\label{alg:apollow}
\end{algorithm}

\noindent Algorithm~\ref{alg:apollow} illustrates the algorithm of \textsc{Apollo} with the standard $L_2$ and the decoupled weight decay.
As \textsc{Apollo} memorizes the update direction of the last iteration $d_t$ to update the diagonal Hessian $B_{t+1}$, the application of decoupled weight decay to \textsc{Apollo} is slightly different from AdamW.
The weight decay term is added to the update direction $d_t$, instead of directly to the update of parameters.
We conducted experiments to evaluate \textsc{Apollo} with decoupled weight decay on image classification tasks. The results are provided in Appendix~\ref{appendix:results}.

\section{Experimental Details}
\label{appendix:details}

\begin{table}[t]
\caption{Hyper-parameters of each optimizer on CIFAR-10 and ImageNet.}
\label{tab:appendix:hyp}
\centering
\begin{tabular}[t]{l|c|c}
\toprule
 & \textbf{CIFAR-10} & \textbf{ImageNet} \\
\midrule
SGD         & $\eta=0.1, \,\, \gamma=5e^{-4}$,  & $\eta=0.1, \,\, \gamma=1e^{-4}$ \\
\midrule
Adam        & $\eta=0.001, \,\, \gamma=2.5e^{-1}, \,\, \epsilon=1e^{-8}$ & $\eta=0.001, \,\, \gamma=1e^{-1}, \,\, \epsilon=1e^{-8}$ \\
RAdam       & $\eta=0.001, \,\, \gamma=2.5e^{-1}, \,\, \epsilon=1e^{-8}$ & $\eta=0.001, \,\, \gamma=1e^{-1}, \,\, \epsilon=1e^{-8}$ \\
AdaBelief.  & $\eta=0.001, \,\, \gamma=2.5e^{-1}, \,\, \epsilon=1e^{-8}$ & $\eta=0.001, \,\, \gamma=1e^{-1}, \,\, \epsilon=1e^{-8}$ \\
\midrule
AdaHessian  & $\eta=0.15, \,\, \gamma=1e^{-3}, \,\, \epsilon=1e^{-2}$ & --- \\
\midrule
\textsc{Apollo} & $\eta=0.01, \,\, \gamma=2.5e^{-4}, \,\, \epsilon=1e^{-4}$,  & $\eta=0.01, \,\, \gamma=1e^{-4}, \,\, \epsilon=1e^{-4}$  \\
\textsc{ApolloW} & $\eta=0.01, \,\, \gamma=2.5e^{-2}, \,\, \epsilon=1e^{-4}$,  & $\eta=0.01, \,\, \gamma=1e^{-2}, \,\, \epsilon=1e^{-4}$  \\
\bottomrule
\end{tabular}
\end{table}

\subsection{Image Classification}
\label{appendix:classification}
\paragraph{CIFAR-10}
For CIFAR-10 dataset, we use the ResNet-110 architecture in the public implementation\footnote{\url{https://github.com/bearpaw/pytorch-classification}}.
Note that ResNet-110 is a modified version of ResNet-18~\citep{he2016deep} to adapt the small image size $32\times32$ in CIFAR-10, and is much smaller than standard ResNet-18. 
The number of parameters for ResNet-110 and ResNet-18 are $1.73\,$M and $11.69\,$M, respectively.
The implementation of AdaHessian is based on the public implementation\footnote{\url{https://github.com/davda54/ada-hessian}}.
The training batch size is set to 128.
For each optimizer, we used two learning rate decay strategies.
First, we train the model on CIFAR-10 for 164 epochs and decay the learning rate at the end of 80-th and 120-th epochs by $0.1$.
Second, we also used the cosine annealing schedule~\citep{loshchilov2017sgdr}.
For the cosine annealing schedule, we train a CIFAR-10 model for 200 epochs.

For every optimizer, we comprehensively tuned its hyper-parameters and selected the set of hyper-parameters with the optimal classification accuracy.
Concretely, for SGD, we fixed momentum at $0.9$ and perform grid search of learning rate $\eta \in \{0.05, \,\, 0.1, \,\, 0.2, \,\, 0.5\}$, weight decay rate $\gamma \in [1e^{-4}, \,\, 1e^{-3}]$ with step $1e^{-4}$.
For Adam and RAdam, we fixed $\beta_1=0.9, \,\, \beta_2=0.999, \,\, \epsilon=1e^{-8}$ and grid search learning rate $\eta \in \{1e^{-4}, \,\, 5e^{-4}, \,\, 1e^{-3}, \,\, 5e^{-3}, \,\, 1e^{-2}\}$, weight decay rate $\gamma \in [1e^{-2}, \,\, 5e^{-1}]$ with step $1e^{-2}$.
For AdaBelief, we fixed $\beta_1=0.9, \,\, \beta_2=0.999$, and grid search learning rate $\eta \in \{1e^{-4}, \,\, 5e^{-4}, \,\, 1e^{-3}, \,\, 5e^{-3}, \,\, 1e^{-2}\}$, and $\epsilon \in \{1e^{-6}, \,\, 1e^{-8}, \,\, 1e^{-12}\}$. 
For weight decay, we tried both the standard $L_2$ and the decoupled version of weight decay. 
For $L_2$, we search weight decay rate $\gamma \in [1e^{-4}, \,\, 1e^{-3}]$ with step $1e^{-4}$, and for decoupled version we search weight decay rate $\gamma \in [1e^{-2}, \,\, 5e^{-1}]$ with step $1e^{-2}$. 
For AdaHessian, we fixed $\beta_1=0.9, \,\, \beta_2=0.999$ and grid search $\eta \in \{0.05, \,\, 0.1, \,\, 0.15, \,\, 0.2\}$, weight decay rate $\gamma \in [5e^{-4}, \,\, 5e^{-3}]$ with step $5e^{-4}$ and $\epsilon \in \{1e^{-2}, \,\, 1e^{-4}, \,\, 1e^{-6}\}$.
For \textsc{Apollo}, we fixed $\beta=0.9, \,\, \epsilon=1e^{-4}$ and grid search learning rate $\eta \in \{0.001, \,\, 0.005, \,\, 0.01, \,\,  0.02\}$, weight decay rate $\gamma \in [5e^{-5}, \,\, 1e^{-3}]$ with step $5e^{-5}$.
We explored applying learning rate warmup to all the optimizers and found that \textsc{Apollo} and AdaHessian significantly benefit from warmup.
The impact of warmup on other optimizers is marginal.
Thus, for \textsc{Apollo} and AdaHessian, learning rates are warmed up linearly in the first $500$ updates.
The selected optimal hyper-parameters for each optimizer are summarized in Table~\ref{tab:appendix:hyp}
Random cropping and random horizontal flipping are applied to training data.
For each experiment, we conduct training on one NVIDIA Tesla V100 GPU.

\paragraph{ImageNet}
For ImageNet, we used the neural architecture of ResNeXt-50~\citep{xie2017aggregated}, with training batch size 256.
For each optimizer, we also used the two learning rate decay strategies --- milestone and cosine.
For milestone decay, we train the model for 120 epochs and decay the learning rate at at the end of 40-th and 80-th epochs by $0.1$.
For cosine annealing, we also train each model for 120 epochs with the cosine annealing schedule.
For each optimizer, we fixed all the hyper-parameters selected from CIFAR-10 experiments, except the rate of weight decay $\gamma$ which is tuned on the classification accuracy.
Random cropping and random horizontal flipping are applied to training data.
For each experiment, we conduct training on eight NVIDIA Tesla V100 GPUs.

\subsection{Language Modeling}
\label{appendix:lm}
One Billion Words dataset~\citep{chelba2013one} is a publicly available benchmark for measuring progress of language modeling.
It contains about 0.8 billion tokens with a vocabulary of 793,471 words, including sentence boundary markers.
Different from \citet{liu2020variance} which shrinks the vocabulary to about 0.64 million words, we used the standard vocabulary\footnote{\url{https://github.com/rafaljozefowicz/lm/blob/master/1b_word_vocab.txt}}.
For the language model, we used two-layer LSTM with 2048 hidden states with adaptive softmax and 300-dimensional word embeddings as input.
The cut-offs of the adaptive softmax are set to $[60000, \, 100000, \, 640000]$, which is different from \citet{liu2020variance}.
Dropout~\citep{srivastava2014dropout} is applied to each layer with drop rate of $0.1$.
No weight decay is applied to these optimizers.
Gradient clips with 1.0 are applied to all the optimization methods.

For each optimizer, we comprehensively tuned its learning rate.
Concretely, for SGD, we searched the learning rate $\eta \in \{ 0.05, \, 0.1, \, 0.5, \, 1.0\}$ and $\eta=0.5$ was selected.
For Adam, RAdam and AdaBelief, we fixed $\beta_1=0.9, \,\, \beta_2=0.999$, and searched for $\eta \in \{ 5e^{-4}, \, 1e^{-3}, \, 2e^{-3}, \, 5e^{-3}\}$, and finally $\eta=1e^{-3}$ was selected.
In addition, following \citet{zhuang2020adabelief}, we also tuned $\epsilon$ for AdaBelief (for Adam and RAdam, we fixed $\epsilon=1e^{-8}$). 
We searched $\epsilon \in \{ 1e^{-8}, \, 1e^{-12}, \, 1e^{-16} \}$ and found that $\epsilon=1e^{-12}$ worked best. 
It should be noticed that AdaBelief is very sensitive to the value of $\epsilon$. 
The result in Table~\ref{tab:lm} is obtained using $\epsilon=1e^{-12}$. 
With other values, e.g. $1e^{-8}$ or $1e^{-16}$, the PPL points of AdaBelief are even higher than Adam and RAdam.
Thus, we suspected that the improvement of AdaBelief over Adam or RAdam on LSTM mainly comes from the fine-tuning of $\epsilon$.
Similar observations were also found in our experiments of image classification, and were reported in \citet{yuan2020eadam}.
For \textsc{Apollo}, we fixed $\beta=0.9, \epsilon=1e^{-4}$, and searched $\eta \in \{0.01, \, 0.05, \, 0.1 \}$.
Finally, $\eta=0.1$ was selected.
Each model is trained for 20 epochs, and the learning rate decays at the end of the 12-th and 18-th epochs by decay rate $0.1$.
LSTMs are unrolled for 20 steps without resetting the LSTM states and the batch size is set to 128. 
Every models is trained on one NVIDIA Titan RTX GPU.

\subsection{Neural Machine Translation}
\label{appendix:nmt}
Our experiments on WMT 2014 English-German are based on the Transformer-base model \citep{vaswani2017attention}, with implementation from the FairSeq package~\citep{ott2019fairseq}.
This dataset contains 4.5M parallel sentence pairs for training.
We following the standard setting~\citep{vaswani2017attention}, using Newstest2013 as the validation set and  Newstest2014 as the test set.
The dataset is pre-processed following \citep{flowseq2019}, using the scripts from FairSeq package\footnote{\url{https://github.com/pytorch/fairseq}}.
Specifically, we use word embedding with 512 dimension and 6-layer encoder/decoder with 8 multi-head attention and 2048 feed-forward dimensions. 
We apply 0.1 label smoothing~\citep{szegedy2016rethinking}, and perform totally $500,000$ updates to train each model.
For Adam, RAdam and AdaBelief, we use start learning rate $0.0005$.
For Adam we set $\beta=(0.9, 0.98)$, while for RAdam and AdaBelief we set $\beta=(0.9, 0.999)$.
For SGD and \textsc{Apollo}, the start learning rates is $0.1$.
The momentum of SGD is $0.9$.
For learning rate scheduling, we applied linear warm up the learning rate for SGD, Adam, AdaBelief, and  \textsc{Apollo} ---  $4000$ updates for Adam and $1000$ updates for SGD, AdaBelief and \textsc{Apollo}.
For RAdam, we did not apply warm up because RAdam is inherently designed to avoid it.
After learning rate warming up, we applied the inverse square root decay~\citep{vaswani2017attention} to Adam.
For SGD, RAdam, AdaBelief and \textsc{Apollo}, we decayed the learning rate at the $300,000$ and $450,000$ updates by decay rate 0.1.
Gradient clips with 1.0 are applied to all the optimization methods, and the dropout ratio are set to $0.1$.
Weight decay rates are $1e^{-4}$ for Adam-type methods, $1e^{-6}$ for SGD, and $1e^{-8}$ for \textsc{Apollo}.
The decoding beam size is set to 5, and the checkpoints of the last 10 epochs are averaged before evaluation.
For each experiment, we conducted distributed training across eight NVIDIA Tesla V100 GPUs with maximum batch size as 8192 tokens per GPU (totally $8192\times 8$ tokens per batch).

\subsection{The Choice of $\sigma$}
\label{appendix:sigma}
In our final version, we change $\sigma$ from 1.0 to 0.01 to make the learning rate $\eta$ of \textsc{Apollo} in a suitable range.
Concretely, in the case of $\sigma=1.0$, the optimal $\eta$ for image classification, language modeling and machine translation are $1.0$, $10.0$ and $10.0$, respectively. 
These values are very different from previous algorithms.
After we changed $\sigma=0.01$, the optimal $\eta$ for the three tasks because $0.01$, $0.1$ and $0.1$, which are in a more acceptable range.
Note that we change $sigma=0.01$ only for the consideration of the convenient application of \textsc{Apollo}. 
It has no affect on the behavior of the algorithm.

\section{Detailed Experimental Results}
\label{appendix:results}
In this section, we report the detailed experimental results in Section~\ref{sec:experiments}, and the results of investigation of the effect of weight decay.

\begin{figure}[t]
\centering
\includegraphics[width=1.0\textwidth]{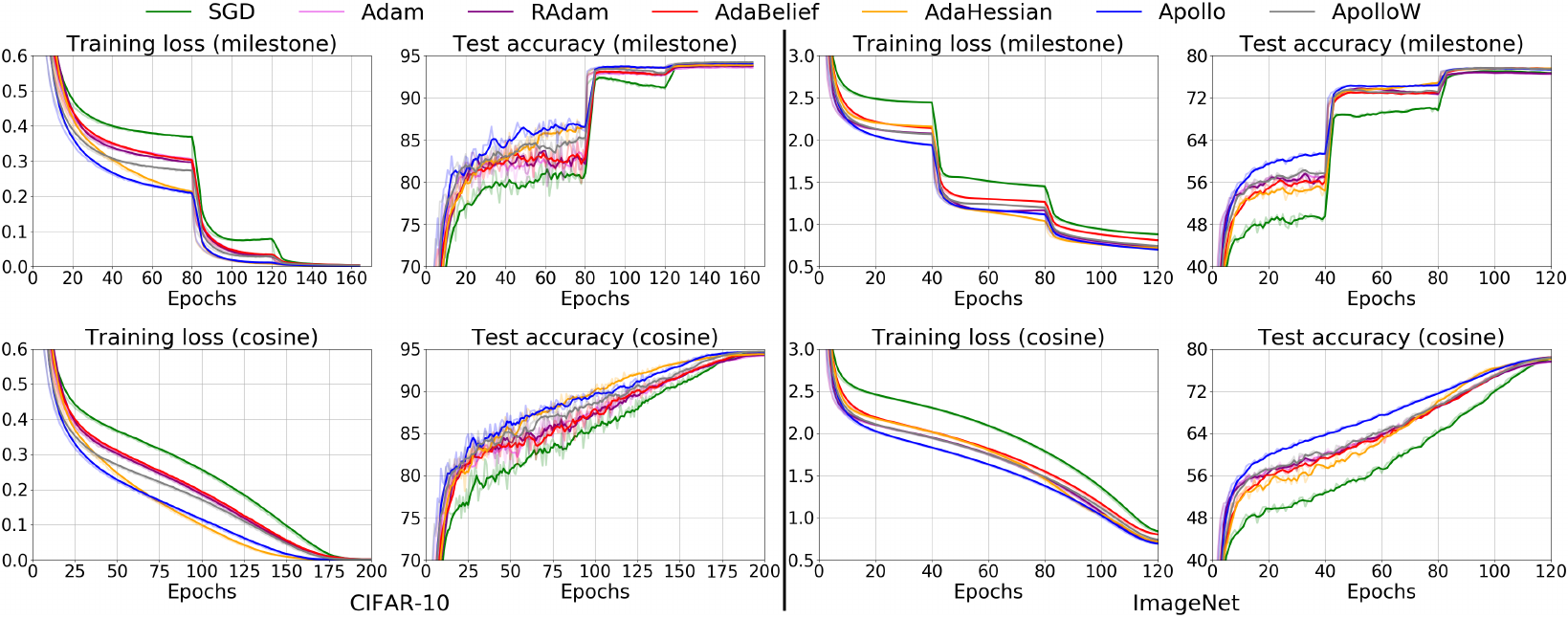}
\caption{Training loss and test accuracy of ResNet-110 on CIFAR-10 and ResNeXt-50 on ImageNet, with two schedule strategies of learning rate decay.}
\label{fig:appendix:classify}
\end{figure}

\begin{table}[t]
\caption{Classification accuracy on CIFAR-10 and ImageNet. For each experiment, we report the mean and standard variance over 5 runs.}
\label{tab:appendix:classify}
\centering
{\small
\begin{tabular}[t]{l|cc|cc}
\toprule
 & \multicolumn{2}{c|}{\textbf{CIFAR-10}} & \multicolumn{2}{c}{\textbf{ImageNet}} \\
\textbf{Method} & \textbf{milestone decay} & \textbf{cosine annealing} & \textbf{milestone decay} & \textbf{cosine annealing} \\
\midrule
SGD         & 93.94$\pm 0.07$ & 94.53$\pm 0.27$ & 77.57$\pm0.07$ & 78.26$\pm0.08$ \\
\midrule
Adam$^*$    & 91.41$\pm0.30$ & 91.56$\pm0.19$ & 71.72$\pm0.13$ & 71.19$\pm0.10$ \\
RAdam$^*$   & 91.80$\pm0.04$ & 91.88$\pm0.15$ & 72.37$\pm0.08$ & 71.64$\pm0.14$ \\
\midrule
Adam        & 93.74$\pm0.15$ & 94.24$\pm0.09$ & 76.86$\pm0.06$ & 77.54$\pm0.16$ \\
RAdam       & 93.88$\pm0.11$ & 94.38$\pm0.25$ & 76.91$\pm0.07$ & 77.68$\pm0.08$ \\
AdaBelief   & 94.03$\pm0.11$ & 94.51$\pm0.07$ & 77.55$\pm0.08$ & 78.22$\pm0.11$ \\
\midrule
AdaHessian  & 93.97$\pm0.22$ & 94.48$\pm0.17$ &  77.61$\pm0.09$ & 78.02$\pm0.10$ \\
\midrule
\textsc{Apollo} & \textbf{94.21}$\pm0.08$ & \textbf{94.64}$\pm0.09$ & \textbf{77.85}$\pm0.07$ & \textbf{78.45}$\pm0.06$ \\
\textsc{ApolloW} & \textbf{94.34}$\pm0.12$ & \textbf{94.76}$\pm0.07$ & \textbf{77.86}$\pm0.09$ & \textbf{78.48}$\pm0.07$ \\
\bottomrule
\end{tabular}
}
\end{table}

\subsection{Detailed Results on Image Classification}
Figure~\ref{fig:appendix:classify} and Table~\ref{tab:appendix:classify} illustrate the details of the experimental results on Image Classification.
For each experiment, we report the mean values with corresponding standard deviations over 5 runs.
Though \citet{loshchilov2019decoupled} claimed that the optimal settings of the learning rate and weight decay factor in Adam with decoupled weight decay is more independent than the original Adam, we observed that the strength of weight decay regularization is still co-related with the learning rate.
To illustrate the significant effect of weight decay strength on both the performance of convergence and generalization, 
we also report the performance of Adam and RAdam with the same weight decay rate of SGD, named Adam$^*$ and RAdam$^*$.

From Figure~\ref{fig:appendix:classify} and Table~\ref{tab:appendix:classify}, we see that Adam$^*$ and RAdam$^*$, with the same weight decay rate of SGD, converge much faster than other optimization methods, while obtaining significantly worse classification accuracy.
After adjusting the weight decay rates, the test accuracy of Adam and RAdam remarkably improves, with rapid decline of convergence speed.
This suggests that the fast convergence speed of Adam and RAdam results from relatively weak regularization.
Thus, the effect of regularization strength needs to be considered when we analyze the performance of different optimization methods.

In addition, we also report the results of \textsc{Apollo} with decoupled weight decay, which is denoted as \textsc{ApolloW}. 
The hyper-parameters of \textsc{ApolloW} (see Table~\ref{tab:appendix:hyp}) are exactly the same of the optimal ones of \textsc{Apollo}.
From Figure~\ref{fig:appendix:classify} and Table~\ref{tab:appendix:classify}, we see that 
\textsc{Apollo} with the standard $L_2$ regularization achieves faster convergence speed, while \textsc{ApolloW} with the decoupled weight decay achieves slightly better generalization accuracy.
Importantly, comparing with Adam-type methods whose performance is significantly impacted by different weight decay implementations, \textsc{Apollo} is much more consistent with the two implementations of weight decay.

\subsection{Effect of Weight Decay Rate on Optimization}
\label{appendix:weight_decay}
To further investigate the effect of weight decay rate on converge speed and generalization performance for different optimization methods, we conduct experiments on CIFAR-10 of ResNet-110 with a range of weight decay rates.
Concretely, we use the weight decay rates $\gamma$ in Table~\ref{tab:appendix:hyp} as the base, and explore different $\gamma$ that are $\alpha$ times of the base weight decay rate, with $\alpha \in \{ 0.2, 0.6, 1.0, 1.4, 1.8\}$.

\begin{figure}[t]
\centering
\includegraphics[width=1.0\textwidth]{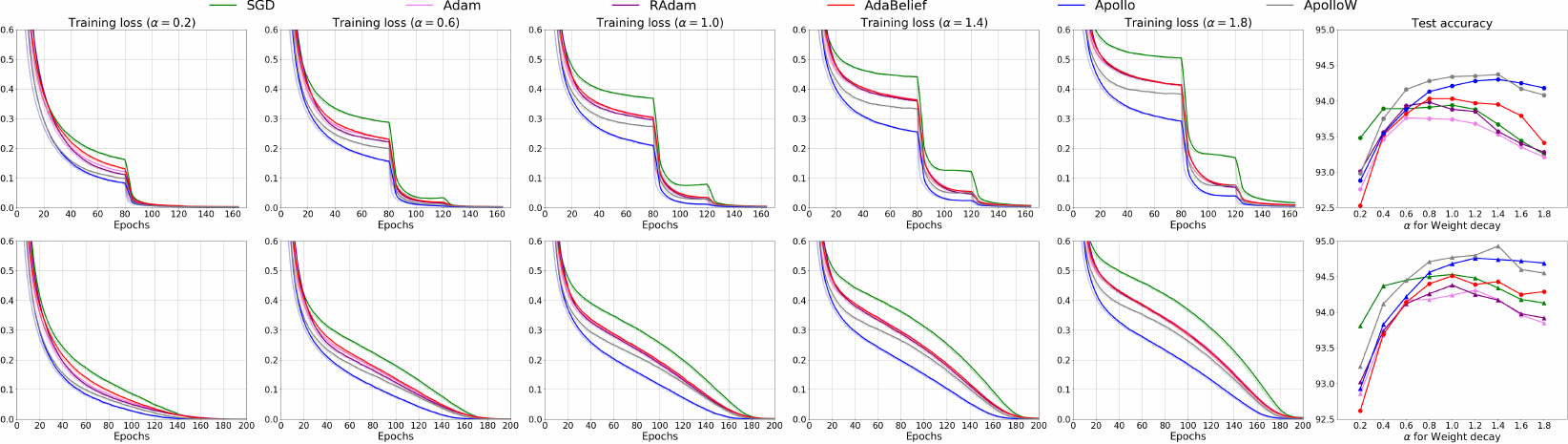}
\caption{Training loss and test accuracy of ResNet-110 on CIFAR-10 with various rates of weight decay, with two schedule strategies of learning rate decay.}
\label{fig:weight_decay}
\end{figure}

Figure~\ref{fig:weight_decay} shows the convergence of different optimization methods with various rates of weight decay, together with the classification accuracy.
\textsc{Apollo} achieves improvements over all the four baselines on convergence speed with different rates of weight decay.
For classification accuracy, \textsc{Apollo} obtains the best accuracy when the weight decay rate ratio $\alpha$ is larger than $0.3$.
When the weight decay rate is decreasing, SGD obtains the best accuracy, while \textsc{Apollo} achieves comparable performance.

\subsection{Comparison on Training Speed and Memory Cost}
\label{appendix:cost}
In this section, we compare the training speed and memory between SGD, Adam, AdaHessian and \textsc{Apollo}.
Table~\ref{tab:cost_compare} summarizes the comparison of cost of a single iteration of update. 
Note that the cost measured in our experiments includes all aspects of model training, including the forward and backward pass of DNNs, not only that of updating parameters for an optimizer.
For fair comparison, experiments of CIFAR-10 and One Billion Words are conduced on a single NVIDIA TITAN RTX GPU, while experiments of ImageNet and WMT are performed with distributed training on 8 NVIDIA Tesla V100 GPUs.

\begin{table}[h]
\centering
\caption{Comparison between different optimization methods on training speed and memory cost. 
Cost includes all aspects of model training, not only that of an optimizer.
}
\label{tab:cost_compare}
{\small
\begin{tabular}{l|cc|cc|cc|cc}
\toprule
 & \multicolumn{2}{c|}{\textbf{CIFAR-10}} & \multicolumn{2}{c|}{\textbf{ImageNet}} & \multicolumn{2}{c|}{\textbf{1BW}} & \multicolumn{2}{c}{\textbf{WMT-14}} \\
Cost ($\times$SGD) & Speed & Memory & Speed & Memory & Speed & Memory & Speed & Memory \\
\midrule
SGD        & 1.00 & 1.00 & 1.00 & 1.00 & 1.00 & 1.00 & 1.00 & 1.00 \\
Adam       & 1.16 & 1.01 & 1.01 & 1.03 & 1.19 & 1.34 & 1.13 & 1.04 \\
Apollo     & 1.42 & 1.01 & 1.23 & 1.05 & 1.49 & 1.62 & 1.19 & 1.06 \\
AdaHessian & 5.76 & 2.12 & 11.78 & 2.51 & 3.51 & 2.78 & 8.46 & 2.47 \\
\bottomrule
\end{tabular}
}
\end{table}

\noindent From Table~\ref{tab:cost_compare}, we see that the second-order AdaHessian requires much more computational resource than first-order methods on both time and memory.
In addition, the slow-down of AdaHessian becomes more significant for larger-scale models with distributed training across multiple GPUS, such as ResNext-50 on ImageNet and Transformer on WMT.

\subsection{Experiments on Parameter-Wise Gradient Clipping}
\label{appendix:pclip}
In this section, we provide some preliminary results on parameter-wise gradient clipping, a modification of  the standard gradient clipping that is inherently proper to \textsc{Apollo}. 
Parameter-wise gradient clipping is to clip the gradient of each parameter individually based on its own norm. 
It can be regarded as a trade-off between gradient clipping by global norm and by each value.

We conducted two groups of experiments to compare with the standard gradient clipping method --- language modeling and neural machine translation.
The experimental settings for standard gradient clipping are exactly the same as in section~\ref{sec:experiments}, where we clipped the gradient by global norm $1.0$ for each model.
For parameter-wise gradient clipping, we clipped each parameter by $0.5$ for the LSTM model in language modeling, and $0.1$ for the Transformer-base model in NMT.
\begin{table}[h]
\caption{Comparison between \textsc{Apollo} with standard and parameter-wise gradient clipping on One Billion Words and WMT-14. We report the mean and standard variance over 5 runs.}
\label{tab:appendix:pclip}
\centering
\begin{tabular}[t]{l|cc}
\toprule
 & \textbf{1BW} & \textbf{WMT-14} \\
\midrule
Standard & 31.94$\pm0.09$ & 28.34$\pm0.10$ \\
Parameter-wise & \textbf{31.75}$\pm0.10$ & \textbf{28.39}$\pm0.11$ \\
\bottomrule
\end{tabular}
\end{table}

\noindent Table~\ref{tab:appendix:pclip} lists the preliminary results. 
On both the two groups of experiments, parameter-wise gradient clipping slightly outperforms the standard one.

\section{Experiments with Small Toy CNN Models}
\label{appendix:toy}
In this section, we provide the comparison between SdLBFGS~\citep{wang2017stochastic} and \textsc{Apollo} on CIFAR-10 dataset with a small toy CNN model\footnote{\url{https://pytorch.org/tutorials/beginner/blitz/cifar10_tutorial.html}}.
The implementation of SdLBFGS is based on the public PyTorch release\footnote{\url{https://github.com/harryliew/SdLBFGS}}, which includes two important modifications to the original SdLBFGS algorithm: identity matrix initialization and direction normalization~\citep{li2018implementation}.
For each optimizer, we train the CNN model for 50 epochs with batch size equals to $64$.
After each epoch, the learning rate is decayed by the rate $0.95$.
For the start learning rate for each optimizer, we performed search in a wide range: $\eta \in \{0.2, 0.1, 0.05, 0.01, 0.005, 0.002, 0.001, 0.0005, 0.0002\}$, and select the one obtains the optimal performance.
The final start learning rates for SdLBFGS and \textsc{Apollo} are $0.1$ and $0.001$, respectively.
Following \citet{li2018implementation}, the memory size of SdLBFGS is set to $100$.
For \textsc{Apollo}, we linearly warmed up the learning rate from $0.01$ in the first $10$ updates.
For other hyper-parameters of each optimizer, we choose the default value.

\begin{figure}[h]
\centering
\includegraphics[width=0.7\textwidth]{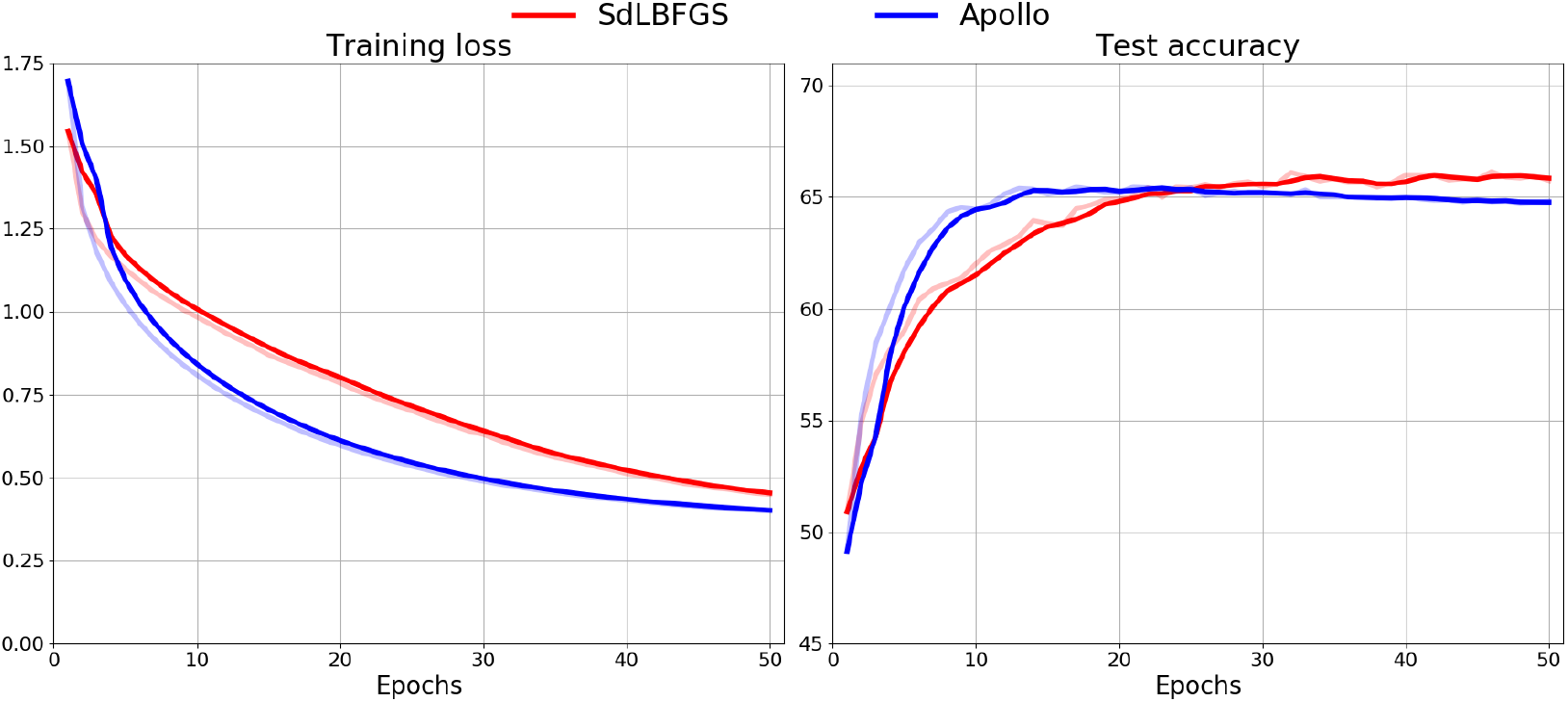}
\caption{Training loss and test accuracy of SdLBFGS and \textsc{Apollo} on CIFAR-10 with the small toy CNN model.}
\label{fig:toy}
\end{figure}

From Figure~\ref{fig:toy}, we see that \textsc{Apollo} convergences faster than SdLBFGS and obtains comparable test accuracy.
Note that \textsc{Apollo} is much faster (more than 10 times for one iteration) than SdLBFGS and consumes much less memory (SdLBFGS stores 100 previous update directions).

\section{Scale-Invariant Apollo}
\label{appendix:scale-invariant}
In \eqref{eq:retify}, we rectify the absolute value of $B_t$ with a convexity hyper-parameter $\sigma$:
\begin{displaymath}
    D_t = \mathrm{rectify}(B_t, \sigma) = \max(|B_t|, \sigma)
\end{displaymath}
To make \textsc{Apollo} scale-invariant, we modify this rectification operation by incorporating a  term similar to the gradient ``belief''~\citep{zhuang2020adabelief}:
\begin{equation}\label{eq:rectify2}
    D_t = \mathrm{rectify}(B_t, \sigma) = \max(|B_t|, \sigma \|g_{t+1} - g_t \|_{\infty})
\end{equation}
It is not hard to prove that \textsc{Apollo} with the rectification in \eqref{eq:rectify2} is scale-invariant.
Importantly, after this modification, $\sigma$ is still coupled with the stepsize $\eta$, and we can set $\sigma=1$ in practice. 
Thus, we do not introduce new hyper-parameters.

\end{document}